\documentclass{article} % For LaTeX2e
\usepackage{iclr2026_conference,times}

\usepackage{amsmath,amsfonts,bm}

\def\eqref#1{equation~\ref{#1}}
\def\1{\bm{1}}

\DeclareMathAlphabet{\mathsfit}{\encodingdefault}{\sfdefault}{m}{sl}
\SetMathAlphabet{\mathsfit}{bold}{\encodingdefault}{\sfdefault}{bx}{n}

\usepackage{hyperref}
\usepackage{url}

\usepackage[utf8]{inputenc} % allow utf-8 input
\usepackage[T1]{fontenc}    % use 8-bit T1 fonts
\usepackage{hyperref}       % hyperlinks
\usepackage{url}            % simple URL typesetting
\usepackage{booktabs}       % professional-quality tables
\usepackage{amsfonts}       % blackboard math symbols
\usepackage{amsmath}        % math environments and symbols
\usepackage{nicefrac}       % compact symbols for 1/2, etc.
\usepackage{microtype}      % microtypography
\usepackage{xcolor}         % colors
\usepackage{graphicx}         % colors
\usepackage{colortbl} 
\usepackage{wrapfig}
\usepackage{enumitem}

\title{The Signal is in the Steps: Local Scoring for Reasoning Data Selection}

\author{Hoang Anh Just \\
Virginia Tech\\
\texttt{just@vt.edu} \\
\And
Myeongseob Ko \\
Virginia Tech \\
\texttt{myeongseob@vt.edu} \\
\And
Ruoxi Jia \\
Virginia Tech \\
\texttt{ruoxijia@vt.edu}
}

\usepackage{amsmath}
\usepackage{amssymb}
\usepackage{mathtools}
\usepackage{amsthm}

\usepackage[utf8]{inputenc} % allow utf-8 input
\usepackage[T1]{fontenc}    % use 8-bit T1 fonts
\usepackage{hyperref}       % hyperlinks
\usepackage{url}            % simple URL typesetting
\usepackage{booktabs}       % professional-quality tables
\usepackage{amsfonts}       % blackboard math symbols
\usepackage{amsmath}        % math environments and symbols
\usepackage{nicefrac}       % compact symbols for 1/2, etc.
\usepackage{microtype}      % microtypography
\usepackage{xcolor}         % colors
\usepackage{graphicx}         % colors
\usepackage{colortbl} 
\usepackage{wrapfig}
\usepackage{enumitem}

\usepackage[capitalize,noabbrev]{cleveref}

\theoremstyle{plain}
\newtheorem{theorem}{Theorem}[section]
\newtheorem{proposition}[theorem]{Proposition}

\newtheorem{corollary}[theorem]{Corollary}
\theoremstyle{definition}

\theoremstyle{remark}

\iclrfinalcopy % Uncomment for camera-ready version, but NOT for submission.
\begin{document}

\maketitle

\begin{abstract}
Distilling long-form reasoning from teacher models into smaller students requires selecting which candidate solutions to train on. Recent work argues that one should select responses the student model assigns highest probability, i.e., favoring solutions ``natural'' to the student. However, we find that this approach works within a single teacher but fails when scaling to long reasoning traces from multiple diverse teachers. We identify a key cause: this approach scores entire solutions, but students generalize by recombining familiar reasoning steps, not by memorizing complete solutions. Full-trajectory scoring optimizes the wrong target; it rewards global fluency while the transferable signal lies in local step transitions. We propose Local Average Log Probability (LALP), which scores each reasoning step using only a small window of preceding context, measuring whether each step is justified by its immediate premises rather than whether the full response looks natural to the student. LALP enables two practical use cases: selecting the best teacher before fine-tuning and curating training data from diverse teacher pools. Across math, coding, and science reasoning tasks, LALP consistently improves accuracy when selecting the most natural solutions by a large margin.

\end{abstract}

\section{Introduction}
\label{sec:intro}

Training large language models (LLMs) to reliably perform multi-step reasoning has become a central challenge. Two complementary paradigms have emerged. Reinforcement learning from verifiable rewards can discover new reasoning behaviors by optimizing task-level outcomes~\citep{Lambert2024TLU3P, Shao2024DeepSeekMathPT}. Supervised distillation from stronger teacher models offers a more direct route to transferring reasoning ability into smaller, deployable students~\citep{Hinton2015DistillingTK}. In practice, distillation can be surprisingly strong: for example, DeepSeek reports a distilled 7B student that surpasses RL-trained 32B models on reasoning benchmarks~\citep{DeepSeekAI2025DeepSeekR1IR}. These results make data curation for distillation a primary lever for improving reasoning capability under tight compute and deployment constraints.

Most existing curation methods focus on prompt selection, i.e., choosing questions to maximize diversity, difficulty, or coverage~\citep{Ash2019DeepBA, Sorscher2022BeyondNS, Albalak2024ASO, Yang2025Select2ReasonEI, Yu2025RethinkingTG}. This line of work treats each prompt as having a single canonical target. Modern reasoning pipelines, however, often have access to multiple teachers (e.g., DeepSeek-R1, QwQ, Qwen3), and each teacher can produce multiple distinct, answer-correct solutions to the same prompt. These candidate responses can differ dramatically in verbosity, stylistic scaffolding, and step structure. This motivates the question we study: \emph{given multiple candidate responses for the same prompt, which response should we train on?}

A natural approach is influence-based selection: estimate each candidate’s downstream impact and train on the most influential traces. However, influence estimation~\cite{xia2024less} typically requires expensive gradient computations, making it impractical at the scale of long-form reasoning data. GRAPE~\citep{zhang2025best} proposes a simple alternative that is both efficient and principled: choose the response that the student model assigns highest probability and select the response that best fits the student’s pretrained distribution. This probability-based criterion yields substantial gains in instruction-following benchmarks where responses are relatively short~\citep{zhang2025best}.

We investigate whether this probability-based selection principle extends to long-form reasoning from heterogeneous teachers. Intuitively, it should: when candidates vary widely in form, selecting what is "natural" to the student seems like it ought to reduce training difficulty and improve outcomes. Interestingly, in the long, mixed-teacher regime, we find that global likelihood produces the wrong ranking: teachers whose traces receive higher probability under the student can yield lower downstream accuracy after fine-tuning. The criterion that works for short, single-teacher responses fails for long, multi-teacher reasoning.

We hypothesize that the key reason is a granularity mismatch between what this likelihood measures and how reasoning generalizes. We analyze student generalization in representation space and find a consistent pattern: test solutions rarely resemble any single training trajectory, while individual reasoning steps are densely covered by the training data. This suggests that students generalize by recombining reusable steps rather than retrieving whole solutions. This suggests selection should evaluate step-level learnability, not trajectory-level fluency.

We therefore propose Local Average Log Probability (LALP), which scores responses at the granularity where reasoning appears to transfer: the step. LALP segments each candidate into reasoning steps, evaluates each step under a small local context window, and averages across steps. This measures whether each transition is locally supported by its immediate premises, rather than whether the full document reads smoothly under long-range conditioning.

LALP enables two practical forms of curation. First, for teacher selection: LALP scores computed before fine-tuning correctly predict which teacher's data yields the best student performance; global likelihood predicts the opposite. Second, for per-prompt response curation from a mixed-teacher pool, LALP improves math accuracy by up to 9.4\% over global likelihood selection. These gains transfer to science reasoning and code generation suggesting that this local scoring rule is a broadly useful principle for reasoning data selection.

\section{Related Work}
\label{sec:related_work}

The task of curating effective data for SFT of LLMs is an active and critical area of research. Our work builds upon and differentiates itself from several existing lines of inquiry, particularly in synthetic reasoning data generation, model-aware data selection, and knowledge distillation.

\paragraph{Synthetic reasoning data.} 
The use of LLMs to generate synthetic chain-of-thought (CoT) responses~\citep{wei2022chain} has become prevalent for tasks requiring multi-step inference~\citep{DeepSeekAI2025DeepSeekR1IR,ye2025limo,liu2025acereason,Bercovich2025LlamaNemotronER}. Recent work has also explored critique-and-revision pipelines using AI evaluators to filter or refine generated examples~\citep{Wu2025BeyondSL, guha2025openthoughts, Chen2025UnveilingTK, Jiang2025WhatMA}. However, as~\citet{guha2025openthoughts} and~\citet{Chandra2025ShapeOT} observed, stronger teachers are not always more beneficial for a given student. Our work addresses the complementary question: when multiple teacher-generated responses are available, how should we select among them?

\paragraph{Model-aware data selection.} 
Recognizing that one-size-fits-all SFT data is suboptimal~\citep{li2025small, Chandra2025ShapeOT}, researchers have explored student-aware selection strategies. GRAPE~\citep{zhang2025best} made a significant step by selecting data based on the student's global average log probability of the entire response, aiming to choose sequences ``natural'' to the student's distribution. Our work directly builds on GRAPE, acknowledging its strengths for single-teacher, short responses but identifying a critical failure mode, the fluency trap, for mixed-teacher settings with long reasoning traces. Other approaches include offline surrogate modeling~\citep{kostrikov2021offline,bai2021pessimistic},   model gradient analysis~\citep{Jung2025PrismaticSG, Panigrahi2025InGG}, online curriculum learning~\citep{liang2021token,lu2021exploiting}, and active learning methods like SIFT~\citep{hubotter2024efficiently}, which combines retrieval and uncertainty reduction. Influence functions have also been explored~\citep{choe2024your, Humane2025InfluenceFF}, though they face scalability challenges for LLMs. Our method offers a simpler, more direct way to achieve model-awareness by operating on the student's inherent probabilities at the step level, without auxiliary models or complex training modifications.

\paragraph{Knowledge distillation.} 
Our approach shares conceptual similarities with knowledge distillation (KD) \cite{Hinton2015DistillingTK}, where knowledge is transferred from a (typically larger) teacher to a smaller student. Classical KD aligns models at the token level by minimizing KL divergence between output distributions at every decoding step~\citep{gou2021knowledge,song2025knowledge}, but this is computationally expensive. A more efficient alternative is response-level KD: the teacher generates complete responses, and the student is trained on those sequences with ordinary cross-entropy loss~\citep{hsieh2023distilling,gupta2023cross, Ding2025MiCoTABT}. Our method belongs to this family but adds a student-aware step-level filter that scores local reasoning transitions rather than entire trajectories.

\section{Problem Setting and Baseline}
\label{sec:problem}

\subsection{Response Selection for SFT}
\label{sec:response-selection}

We study response selection for SFT. Given a prompt $x$ and a candidate set of responses
$\mathcal{Y}(x) = \{y^{(1)}, \ldots, y^{(K)}\}$, generated by sampling multiple teacher models, the goal is to select a single response $y^* \in \mathcal{Y}(x)$ to include in the training set. Throughout the paper, all candidates' final answers match the ground truth. 
Selection therefore focuses not on correctness, but on supervision quality. Which reasoning trace, among several correct ones, provides the most effective training signal for a given student?

Formally, we seek a scoring function $f(y; x, \theta_S)$ that uses the student model (with parameters $\theta_S$) to evaluate candidates, such that selecting
$y^* = \arg\max_{y \in \mathcal{Y}(x)} f(y; x, \theta_S)$
yields the best downstream performance after fine-tuning on the selected data.

\subsection{Baseline: Global Average Log Probability (GALP)}
\label{sec:galp}

A recent approach to this problem is GRAPE \citep{zhang2025best}, which scores each candidate response by its global average log probability under the student model:
\begin{align}
\label{eqn:galp}
    \mathrm{GALP}(y \mid x) &= \frac{1}{m} \sum_{t=1}^{m} \log P_{\theta_S}(y_t \mid y_{<t}, x)\\
    &\propto \log P_{\theta_S}(y|x)
\end{align}
where $m$ is the number of tokens in $y$. 
% Selection chooses the highest-scoring response:
% \begin{equation}
%     y^*_{\mathrm{GALP}} = \arg\max_{y \in \mathcal{Y}(x)} \mathrm{GALP}(y \mid x).
% \end{equation}

GRAPE's hypothesis is that SFT is most effective when training data aligns with the student's pretrained distribution. Responses that are globally ``natural'' to the student, those it assigns high probability, should be easier to learn and lead to better outcomes. This criterion is also computationally attractive: scoring requires only a single forward pass per candidate. In this paper, we stress-test GALP in a regime it was not originally tested on: long reasoning traces (over 12K tokens) drawn from heterogeneous teachers with systematically different verbosity and style.

% GRAPE demonstrated strong gains in instruction-tuning benchmarks, where responses are typically short. In this paper, we test whether GALP remains effective in a different regime: long-form reasoning traces (over 10K tokens) from heterogeneous teachers. 

\section{When Global Selection Fails}
\label{sec:when-galp-fails}

This section presents the empirical motivation for our approach: in long-form reasoning distilled from heterogeneous teachers, selecting by GALP becomes an unreliable proxy for supervision quality. We begin with a controlled within-teacher sanity check where GALP behaves as intended, and then show two breakdowns in the long mixed-teacher regime: a teacher-level ranking reversal and degraded per-prompt selection.

\paragraph{When GALP Works (Sanity Check).} We first verify that GALP is not a strawman. When candidate responses are generated by the same teacher (comparable style and similar lengths), selecting higher-GALP responses improves downstream accuracy. Table~\ref{tab:within_teacher_galp} shows results in this controlled setting. Within each teacher, we partition responses into the lowest, middle, and highest GALP terciles (computed by the pre-SFT student) and fine-tune the student on each subset. Selecting from the highest-GALP tercile consistently outperforms selecting from the lower terciles.

\begin{table}[t]
\centering
\resizebox{0.5\columnwidth}{!}{%  <-- Recommended: remove resizebox for narrow tables, or use 0.5\textwidth
\begin{tabular}{llr}
                                       % &   & \textbf{AVG}  \\ 
                                       \toprule
                                        \multicolumn{2}{c}{\textbf{Student: Qwen2.5-7B-Instruct}} & \textbf{AVG}\\ \toprule
                                       & Original Model           & 0.353 \\
                                       \midrule
\textbf{Teacher:} & Lowest GALP  & 0.292 \\
\textbf{Qwen2.5-72B} & Middle GALP  & 0.338 \\
\textbf{-Instruct}   & Highest GALP & 0.368 \\
                                       \midrule
\textbf{Teacher:} & Lowest GALP  & 0.320 \\
\textbf{QwQ-32B}  & Middle GALP  & 0.349 \\
                  & Highest GALP & 0.382 \\
                                       \bottomrule
\end{tabular}
}
\caption{Within-teacher sanity check for GALP. For each teacher, responses are partitioned into terciles by the pre-SFT student's global average log probability (Eq.~\ref{eqn:galp}). Fine-tuning on the highest-GALP tercile yields the strongest average downstream performance.}
\label{tab:within_teacher_galp}
\end{table}

% The key question is whether this behavior persists when candidates come from heterogeneous sources with dramatically different verbosity and stylistic conventions.

\begin{figure*}[t]
    \centering
    \includegraphics[width=0.65\linewidth]{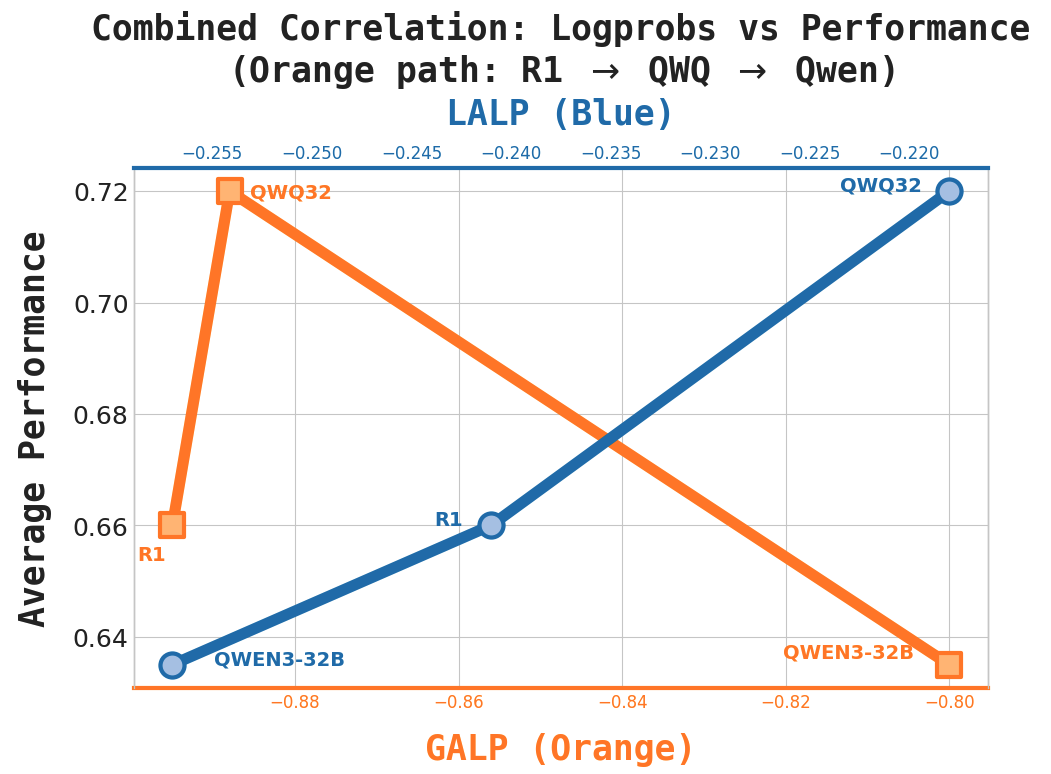}
    \caption{Cross-teacher reversal under global likelihood. For each teacher, we plot downstream accuracy after SFT on that teacher's dataset versus the pre-SFT student's average GALP on the same data (scaled by $10^2$ for readability). Across both students, higher average GALP does not imply higher downstream accuracy.} 
    % \ruoxi{prune this figure and only keep minimal relevant part to the section's narrative. one clear figure works better than mulitple unreadable figures. we only need one figure here and move the one with reduance information to appendix and mention it in the caption }}
    \label{fig:galp_teacher_reversal}
\end{figure*}

\paragraph{Empirical Evidence for Failure.} We now consider long-form reasoning responses from three modern teachers: DeepSeek-R1, Qwen3-32B, and QwQ-32B. These models differ substantially in verbosity, self-reflection patterns, and stylistic scaffolding. For each teacher, we generate answer-verified solutions to the same prompts and fine-tune the same student on each teacher-specific dataset.
Figure~\ref{fig:galp_teacher_reversal} shows that ranking teachers by their average GALP under the student does not predict and can even invert the ranking by downstream accuracy. Specifically, Qwen3-32B traces receive the highest average GALP under both 7B and 32B students, yet SFT on these traces yields the worst downstream accuracy; QwQ-32B traces do not achieve the highest GALP, yet SFT on them yields the best downstream accuracy.

% The cross-teacher reversal shows that GALP is confounded at the teacher level. 
We further test GALP in a more fine-grained setting: per-prompt selection from a mixed-teacher pool. For each prompt, we pool answer-verified responses from all three teachers and select one response per prompt using: random selection and highest GALP.  We then fine-tune the same student on each curated dataset. As shown in Table 
\ref{tab:data_selection_results_acrosss_teacher_models}, in this mixed-teacher setting, GALP selection underperforms random selection for the 32B student and provides only marginal gains for the 7B student. 
% This indicates that GALP is systematically misled by cross-teacher variation in style and verbosity.

% \paragraph{Summary.} In modern reasoning distillation, where long traces are drawn from heterogeneous teachers, global likelihood can rank data in ways that contradict downstream utility. GALP behaves sensibly within a single teacher, but fails across teachers---both at the dataset level (cross-teacher reversals) and at the per-prompt level (degrading selection quality).

\section{Why GALP Fails}
\label{sec:analysis}

GALP scores a candidate solution by the probability of the full response given the prompt, implicitly treating trajectory likelihood as a proxy for supervision quality. In long-form reasoning, we argue this target is misaligned with how students actually generalize. We provide evidence that reasoning transfers at the granularity of steps, not trajectories, and draw implications for how selection should be designed.

\paragraph{Hypothesis: Step-compositional Generalization.} We probe the granularity at which reasoning supervision transfers from training to test time. For each held-out test solution, we embed it at two levels: (i) a single embedding for the full trajectory, and (ii) separate embeddings for individual reasoning steps. We quantify coverage as the cosine similarity to the nearest training-set neighbor at each level. (See Appendix 
\ref{app:details} for step extraction details.) Higher similarity indicates that test-time reasoning is well-supported by training examples at that granularity.

\begin{figure*}[h!]
    \centering
    \includegraphics[width=0.99\linewidth]{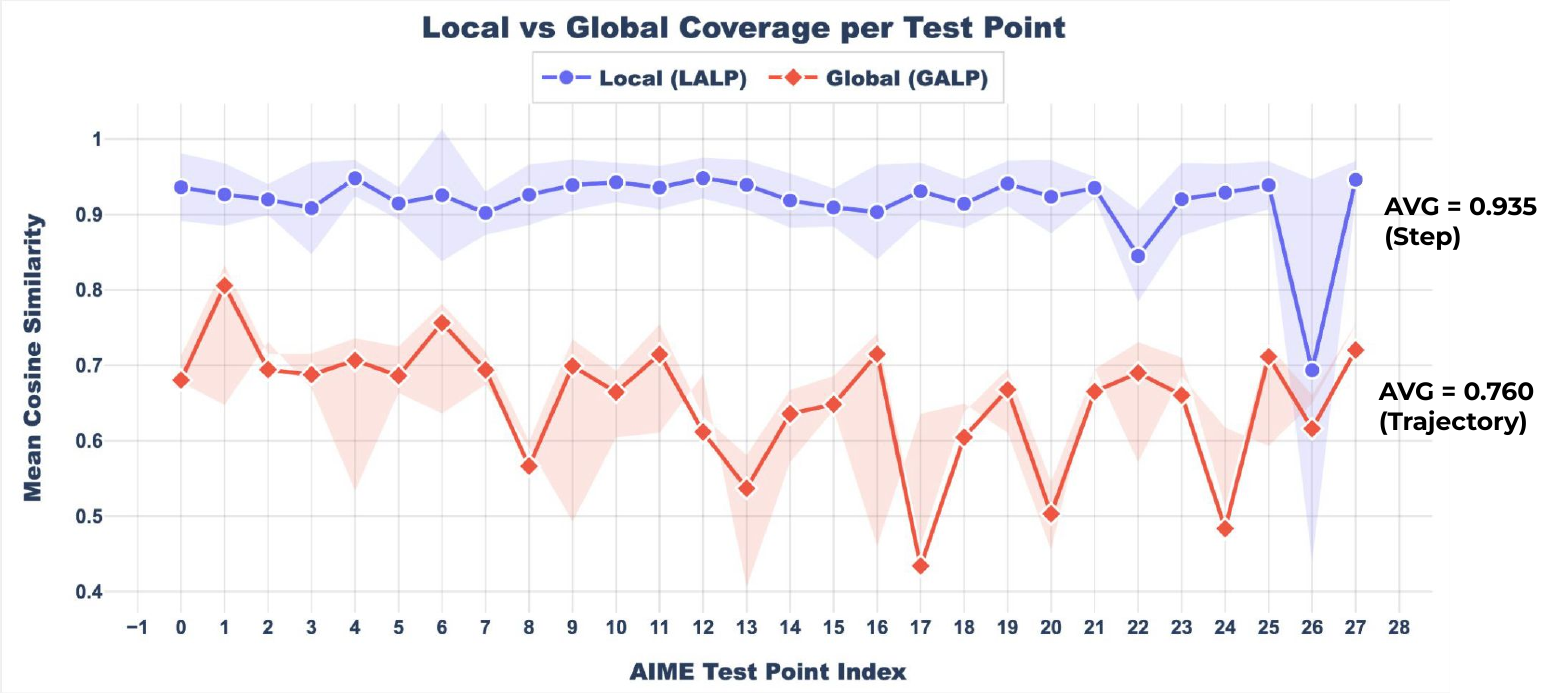}
    \caption{Per-test-point coverage distributions for AIME 2025 problems. Each violin shows the distribution of cosine similarities between a test point and training data. \textbf{Red (Global):} Trajectory-level embeddings show sparse coverage with high variance across test points (mean=0.760, range 0.68--0.83). Many test problems have no close training match. \textbf{Blue (Local):} Step-level embeddings exhibit dense, consistent coverage (mean=0.935, p10=0.89) with tight distributions per test point, allowing reasoning step to find a similar training step. This 23\% coverage gap confirms that step-level representations generalize reliably while full trajectories do not.}
    \label{fig:coverage_global_vs_local}
\end{figure*}

Figure~\ref{fig:coverage_global_vs_local} shows an interesting gap:
step-level embeddings have very high nearest-neighbor similarity (0.935),
while full-trajectory embeddings are substantially less covered (0.760).
This indicates that, at test time, models rarely encounter a training example that matches an entire solution template,
but they likely encounter locally similar reasoning moves. Figure~\ref{fig:coverage_global_vs_local} confirms this gap is consistent across test problems. We observe a similar trend across benchmarks we have tested and provide further results in Appendix \ref{app:test set coverage}.

% \begin{table}[t]
% \centering
% \small
% \begin{tabular}{lcc}
% \toprule
% \textbf{Granularity} & \textbf{Avg.\ 1-NN cosine} & \textbf{\# items} \\
% \midrule
% Trajectory (full response) & 0.760 & 30 \\
% Step (sentence)            & 0.935 & 3{,}437 \\
% \bottomrule
% \end{tabular}
% \caption{Nearest-neighbor similarity between evaluation reasoning and training data at two granularities (1-NN cosine in embedding space). Step-level similarity is substantially higher than trajectory-level similarity, consistent with solutions being novel combinations of locally familiar moves.}
% \label{tab:embedding_coverage}
% \vspace{-0.8em}
% \end{table}

These statistics support a step-compositional view of reasoning: students improve by learning a library of reusable local transitions and recombining them into novel trajectories.
This perspective aligns with the basic structure of autoregressive inference:
at each point in a solution, the model must produce a plausible next step given the prompt and recent progress.
If what transfers is the step, then the learnability of a candidate solution should be governed by the quality of its local transitions, not by the likelihood of the entire response under an arbitrarily long prefix that is largely problem-specific.

\paragraph{Implication for Likelihood-Based Selection.} The coverage gap implies that optimizing for high-probability trajectories is a mismatched target for selection in this regime: full trajectories are novel by design, while their constituent steps are what recur across problems. Selection should instead prioritize solutions whose individual steps are locally natural and reusable for the student. We formalize this intuition in a toy theoretical model (Appendix~\ref{app:toy_theory}), which shows that when a learner has dense support only for local contexts, it acquires signal there but collapses to an uninformative prior for non-local contexts, making global scoring unreliable while local scoring remains informative.

\section{Method: Local Average Log Probability}
\label{sec:method}

The analysis above suggests that likelihood-based selection should (i) condition on the level where reasoning generalizes, steps rather than full context, and (ii) aggregate at the  We formalize this as \textbf{Local Average Log Probability (LALP)}.

\paragraph{Step segmentation.}
LALP requires partitioning a response into a sequence of reasoning steps. We use a capable, open-weight large language model, GLM-4.5-Air~\cite{Zeng2025GLM45AR}, which can handle large context that is necessary for our long reasoning data. Full implementational provided in Appendix~\ref{app: results}.

% Sentence boundaries are an imperfect proxy for reasoning steps: a single mathematical move may span multiple sentences, and a single sentence may contain multiple moves. To assess sensitivity, we evaluated several alternatives---fixed-length chunks (50 tokens), newline-based splits, and manual semantic step annotations for 100 responses. The induced teacher ranking remains stable across these variants (Spearman $\rho > 0.85$; details in Appendix~\ref{app:segmentation}).

\paragraph{Definition.}
Let a response $y$ be segmented into $p$ steps,
$y = (s_1, s_2, \ldots, s_p)$,
where each step $s_i = (s_i^1, s_i^2, \ldots, s_i^{|s_i|})$ is a sequence of $|s_i|$ tokens. We score each step by its token-average log probability under the student model $\theta_S$, conditioning only on the prompt $x$ and the $k$ immediately preceding steps:
% \small{
\begin{equation*}
    \mathrm{LocalLP}(s_i) \;=\; \frac{1}{|s_i|}\sum_{t=1}^{|s_i|}\log P_{\theta_S}\!\bigl(s_i^t \mid s_i^{<t},\, s_{\max(1,i-k):i-1},\, x\bigr),
\end{equation*}
    % }
where $s_i^{<t}$ denotes the tokens within step $i$ before position $t$.
LALP is the average of these step scores:
\begin{equation}
\label{eq:local_log_prob}
    \mathrm{LALP}(y \mid x) \;=\; \frac{1}{p}\sum_{i=1}^{p} \mathrm{LocalLP}(s_i).
\end{equation}
Given multiple candidate responses for a prompt, we select the response with the highest $\mathrm{LALP}(y \mid x)$.

This construction enforces two properties that GALP lacks in the long mixed-teacher regime: (i) equal step weighting (each reasoning move contributes equally, independent of verbosity), and (ii) local conditioning (each step is evaluated for standalone plausibility given its immediate premises, rather than for global document-level fluency).

\begin{figure*}[h!]
    \centering
    \includegraphics[width=0.8\linewidth]{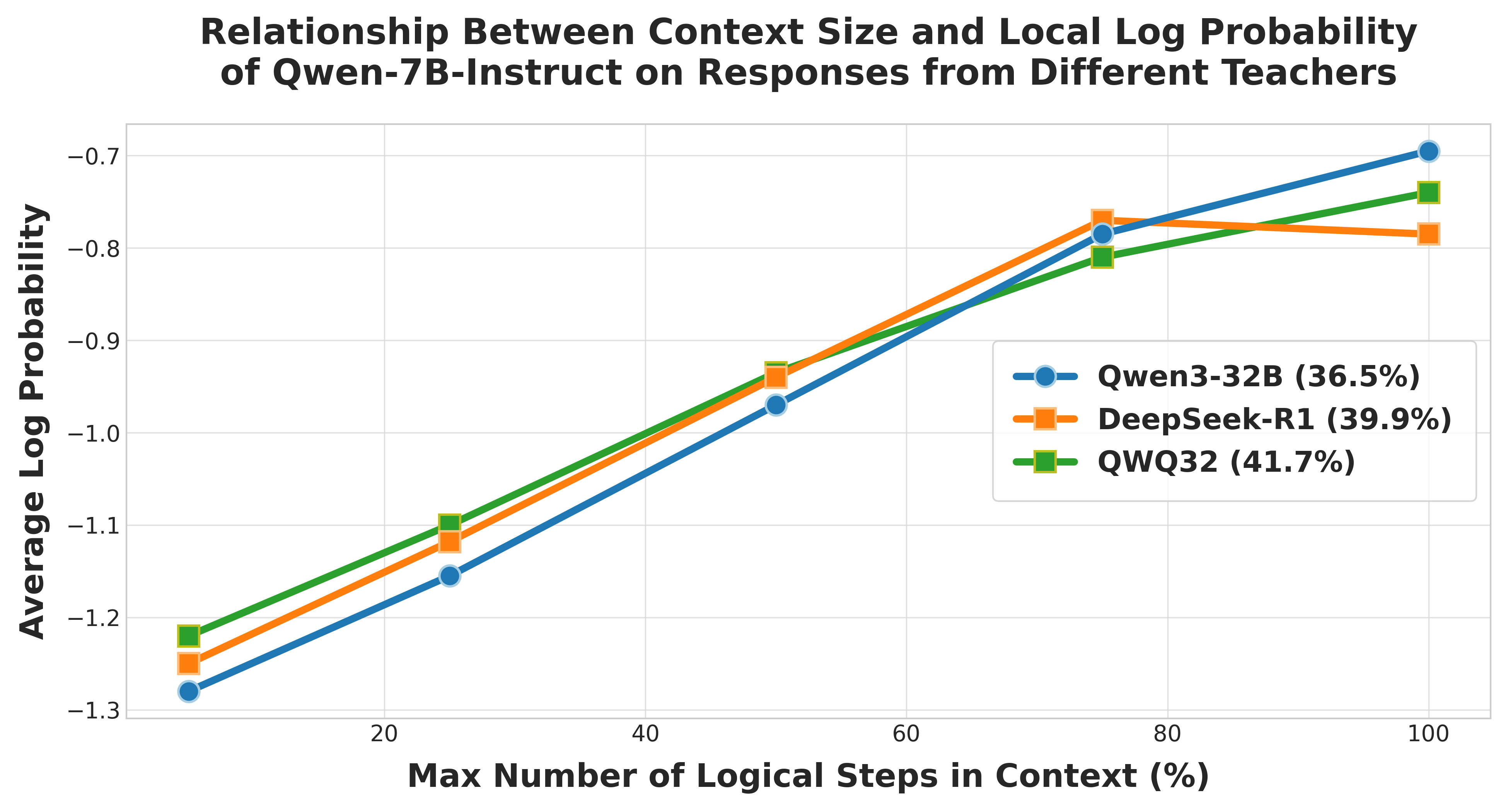}
    \caption{\textbf{Effect of context window on teacher ranking.} Mean step-level log probability (LALP) computed with increasing context windows. Small windows (5--25\%) yield a teacher ordering that matches downstream performance; large windows (50--75\%) converge to the GALP ordering and recover the cross-teacher reversal.}
    \label{fig:context_window_vs_logprobs}
\end{figure*}

\paragraph{Context window size.}
The window size $k$ governs the degree of locality. To study its effect, we set the window size to a fraction $\alpha \in \{5\%, 25\%, 50\%, 75\%\}$ of the available prefix steps for each response, and compare the resulting teacher rankings to the GALP ranking.
Figure~\ref{fig:context_window_vs_logprobs} shows a clear transition: with small windows ($\alpha=5\%$--$25\%$), rankings are stable and match downstream performance (correctly identifying QwQ-32B as the strongest teacher). As $\alpha$ increases, rankings converge toward GALP, eventually recovering the same failure mode where Qwen3-32B ranks highest despite yielding the weakest fine-tuned accuracy.

This ablation indicates that locality is the driver of LALP's advantage and that rankings are robust across a range of small windows. 

% Unless otherwise stated, we use $k=3$ steps.

\paragraph{Practical considerations.}
Computing LALP requires scoring each step under its local window, which entails $p$ forward passes per response (vs.\ a single pass for GALP). In practice, this overhead is manageable for two reasons. First, selection is a one-time preprocessing cost incurred before fine-tuning. Second, for teacher selection, we find that scores computed on as few as 200 prompts reliably recover the full-dataset teacher ranking (Section~\ref{subsec:teacher_selection}). Finally, steps from different responses can be batched to improve GPU utilization.

\begin{table}[h!]
\centering
\resizebox{0.7\columnwidth}{!}{
\begin{tabular}{lccc}
\toprule
\textbf{Model Configuration} & \textbf{OLYMPIAD} & \textbf{CN\_MATH24} & \textbf{AVG} \\
\midrule
\multicolumn{4}{c}{\textbf{Student: Qwen2.5-7B-Instruct}} \\
\midrule
Original Model & 0.404 & 0.167 & 0.353 \\
Random         & 0.456 & 0.233 & 0.407 \\
GALP           & 0.441 & 0.233 & 0.412 \\
Local Lowest   & 0.433 & 0.233 & 0.399 \\
LALP           & 0.441 & 0.333 & \textbf{0.440} \\
\midrule
\multicolumn{4}{c}{\textbf{Student: Qwen2.5-32B-Instruct}} \\
\midrule
Original Model & 0.471 & 0.233 & 0.445 \\
Random         & 0.636 & 0.733 & 0.651 \\
GALP           & 0.636 & 0.733 & 0.632 \\
Local Lowest   & 0.640 & 0.700 & 0.623 \\
LALP           & 0.673 & 0.833 & \textbf{0.726} \\
\bottomrule
\end{tabular}
}
\caption{Teacher selection results showing performance on OlympiadBench, Chinese Math, and the Average over 7 benchmarks. Full results in Table \ref{tab:teacher_selection_results_full}.}
\label{tab:teacher_selection_results}
\end{table}

\section{Evaluation}
\label{sec:experiments}

In this section, we evaluate LALP as a data-selection rule for reasoning distillation in the long, mixed-teacher regime. Our experiments address two practical use cases: (1) selecting the best teacher model before fine-tuning, and (2) per-prompt response selection from a mixed-teacher pool. We find that LALP correctly ranks teachers and improves downstream accuracy by up to 9.4\% over GALP in per-prompt selection. We then analyze three mechanisms underlying these gains: shifted attribution toward reasoning-critical tokens, resistance to self-conditioning artifacts, and a dissociation between training loss and generalization that supports the step-compositional view of reasoning.

\paragraph{Experimental setup.}
We use LIMO prompts~\citep{ye2025limo} for long-form mathematical reasoning.
To test cross-domain transfer, we additionally evaluate on GPQA-Diamond~\citep{rein2023gpqa} (science reasoning) and LiveCodeBench~\citep{Jain2024LiveCodeBenchHA} (code reasoning).
We consider three student models: Qwen2.5-7B-Instruct, Qwen2.5-32B-Instruct, and Llama-3.1-8B-Instruct.
Our teacher pool contains DeepSeek-R1~\citep{DeepSeekAI2025DeepSeekR1IR}, Qwen3-32B-Instruct, and QwQ-32B, which differ substantially in verbosity and stylistic conventions.
We report average accuracy across a diverse suite of math benchmarks (MATH, AIME 2025, AMC, MINERVA, KAOYAN, OlympiadBench, CN\_MATH24); full evaluation details and hyperparameters are provided in Appendix~\ref{app:details}.

\subsection{Teacher Model Selection}
\label{subsec:teacher_selection}

We first test a coarse-grained but practical use case: selecting a teacher before fine-tuning.
Fine-tuning a student on each candidate teacher is expensive; an effective scoring rule should predict the best teacher using only the pre-trained student.

For each teacher, we compute the average GALP and LALP of its responses under the pre-trained student, then fine-tune the student on that teacher's dataset and measure downstream accuracy.
Table~\ref{tab:teacher_selection_results} shows that LALP correctly ranks teachers by downstream utility (QwQ $>$ DeepSeek-R1 $>$ Qwen3) for both 7B and 32B students, while GALP induces the opposite ordering.
In addition, LALP rankings computed from as few as 200 prompts match those from the full 817-prompt set, enabling efficient teacher selection in practice.

\subsection{Per-Prompt Response Selection from a Mixed-Teacher Pool}
\label{subsec:main_result}

\begin{table}[h!]
\centering
\resizebox{0.4\columnwidth}{!}{
\begin{tabular}{lc}
\toprule
\textbf{Model Configuration} & \textbf{AVG} \\
\midrule
\multicolumn{2}{c}{\textbf{Student: Qwen2.5-7B-Instruct}} \\
\midrule
Original Model & 0.353 \\
Random         & 0.407 \\
GALP           & 0.412 \\
Local Lowest   & 0.399 \\
LALP           & \textbf{0.440} \\
\midrule
\multicolumn{2}{c}{\textbf{Student: Qwen2.5-32B-Instruct}} \\
\midrule
Original Model & 0.445 \\
Random         & 0.651 \\
GALP           & 0.632 \\
Local Lowest   & 0.623 \\
LALP           & \textbf{0.726} \\
\bottomrule
\end{tabular}
}
\caption{Response selection from mixed-teacher pool. LALP (Local Highest) outperforms GALP (Global Highest) by \textbf{+9.4\%} on the 32B student (0.726 vs 0.632). Notably, LALP-selected data even outperforms training on all responses from the best single teacher (QwQ-32B: 0.719 in Table~\ref{tab:teacher_selection_results}). Full results in Table~\ref{tab:data_selection_results_acrosss_teacher_models_full}.}
\label{tab:data_selection_results_acrosss_teacher_models}
\end{table}

Teacher selection treats each source as monolithic.
In modern distillation pipelines, practitioners often have multiple candidate traces per prompt (multiple teachers, or multiple samples).
We therefore consider the fine-grained setting: for each prompt, pool correct traces from all teachers and select a single response.

We compare four selection rules:
(1) Random,
(2) GALP (highest global log probability),
(3) LALP (highest local log probability),
and (4) Local Lowest (sanity check).
Students are fine-tuned on the resulting 817-example curated datasets.

Table~\ref{tab:data_selection_results_acrosss_teacher_models} shows that LALP yields the best downstream accuracy for both students, improving by \textbf{+9.4} points over GALP on Qwen2.5-32B-Instruct (0.726 vs.\ 0.632).
Notably, GALP underperforms random selection for the 32B student, indicating that in this regime global likelihood is not merely noisy but systematically misranked.

Finally, the LALP-selected dataset---a mixture drawn from all three teachers---outperforms training on the single best teacher alone (QwQ-32B; Table~\ref{tab:teacher_selection_results}).
This suggests that step-aware selection can exploit complementary strengths across teachers without committing to one source.

\subsection{Generalization Beyond Math}
\label{subsec:generalization}

To test whether LALP’s selection principle extends beyond mathematics, we evaluate models trained on LIMO-selected math data on science (GPQA-Diamond) and code (LiveCodeBench).
Table~\ref{tab:gpqa_results} shows that LALP improves over GALP on both benchmarks (+9.1 points on GPQA-Diamond and +2.9 points on LiveCodeBench-Hard), suggesting the gains are not specific to math-only evaluation.

\subsection{Mechanisms Behind LALP's Gains}
\label{subsec:mechanisms}

The step-transferability hypothesis (Section~\ref{sec:analysis}) explains why trajectory-level scoring is misaligned with reasoning generalization. Here we trace LALP's improvements to three concrete mechanisms that follow from its design.

\subsubsection{Token-level attribution.} 
Long solutions contain two broad classes of tokens:
\begin{itemize}[leftmargin=*,itemsep=2pt]
    \item \textbf{Reasoning-critical tokens}: calculations, bindings, and logical transitions that introduce specific new information and are often locally low-probability.
    \item \textbf{Discourse tokens}: scaffolding (e.g., ``Okay,'' ``Therefore,'' restatements) that is abundant and highly predictable.
\end{itemize}

\begin{figure*}[t]
    \centering
    \includegraphics[width=0.85\linewidth]{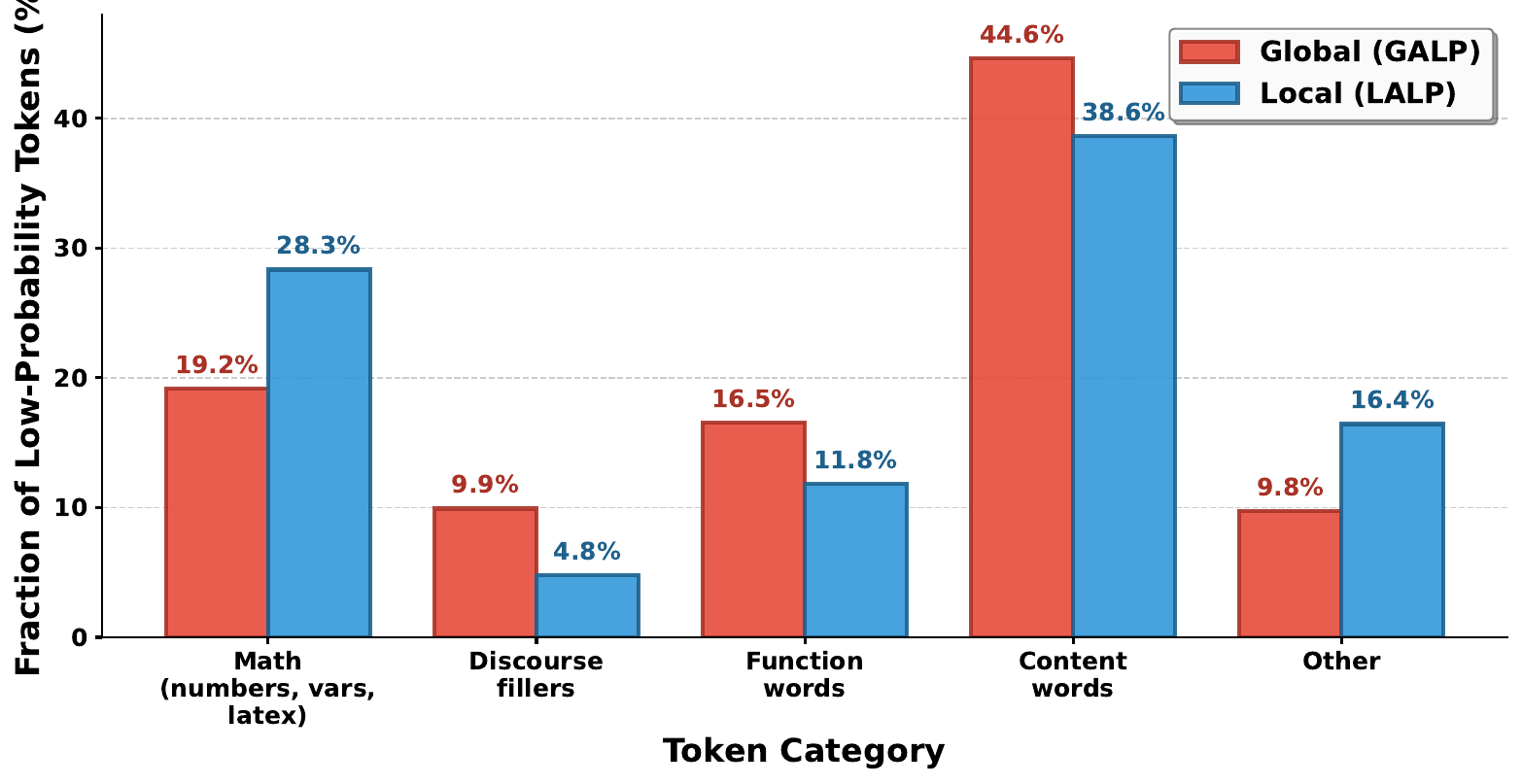}
    \caption{Token type distribution among low-probability tokens under GALP vs.\ LALP. Local scoring identifies a larger fraction of math/symbol tokens among low-probability regions, indicating better focus on reasoning-bearing content. Analysis over 440 LIMO responses across three teachers.}
    \label{fig:token_focus}
\end{figure*}

Because GALP averages over all tokens uniformly, abundant discourse tokens dominate the score by sheer volume, while sparse reasoning-critical tokens contribute minimally. This makes GALP more sensitive to stylistic fluency than to reasoning quality---a phenomenon we formalize as the ``fluency trap'' in Theorem~\ref{thm:signal_dilution_galp} (Appendix~\ref{app:toy_theory}), which shows that when high-probability style tokens outnumber low-probability reasoning tokens, GALP can systematically prefer incorrect responses. We quantify this effect by categorizing tokens in 500 long responses and measuring each method's attribution mass per category. In particular, to measure where each method focuses its ``attention,'' we compute the \emph{probability mass} attributed to each token category. Specifically, for each token $t$ with log probability $\log p_t$, we convert to probability $p_t = \exp(\log p_t)$ and compute the fraction of total probability mass in each category: $\text{mass}_c = \sum_{t \in \text{category } c} p_t \,/\, \sum_{t} p_t$. High-probability tokens (e.g., discourse fillers) naturally dominate GALP's mass because they contribute disproportionately to the sum. For LALP, the mass is computed analogously using local probabilities conditioned on each split's prefix. As shown in Table~\ref{tab:token_category_mass}, GALP assigns 42\% of its mass to discourse tokens, while math/symbols and reasoning transitions together receive only 44\%. 
% \ruoxi{exmain how these percentage of mass is calculated? percentage of log prob or just prob?} 
LALP reverses this balance: discourse tokens account for just 19\%, while math and reasoning tokens together receive 71\%—a shift of 27 percentage points toward reasoning-bearing content. Figure~\ref{fig:token_focus} reveals a complementary pattern among low-probability tokens, which are most informative for distinguishing candidates under likelihood-based scoring. LALP surfaces substantially more math/symbol tokens in this critical region than GALP, indicating that step-level aggregation preserves signal from the sparse logical transitions that token-averaging dilutes. Theorem~\ref{thm:lalp_robustness} formalizes this robustness: LALP's separation between correct and incorrect responses dilutes with the number of \emph{steps} $p$, not the number of \emph{tokens} $m$, providing greater resilience to verbose scaffolding.

\begin{table}[t]
    \centering
    \small
    \resizebox{\columnwidth}{!}{
    \begin{tabular}{lcc}
        \toprule
        \textbf{Token category} & \textbf{GALP (mass \%)} & \textbf{LALP (mass \%)} \\
        \midrule
        Discourse / filler (e.g., ``Okay'', ``So'') & 42.3 & 18.7 \\
        Math / symbols (e.g., variables, operators) & 31.2 & 48.6 \\
        Reasoning transitions (e.g., ``therefore'') & 12.8 & 22.4 \\
        Other & 13.7 & 10.3 \\
        \bottomrule
    \end{tabular}
    }
    \caption{Token-level signal analysis via attribution mass. Under GALP, a large fraction of weight falls on predictable discourse tokens, while LALP shifts emphasis toward math/symbol tokens and reasoning transitions.}
    \label{tab:token_category_mass}
    \vspace{-0.8em}
\end{table}

\subsubsection{Local conditioning resists self-reinforcement.} 
\begin{figure*}[h!]
    \centering
    \includegraphics[width=0.69\linewidth]{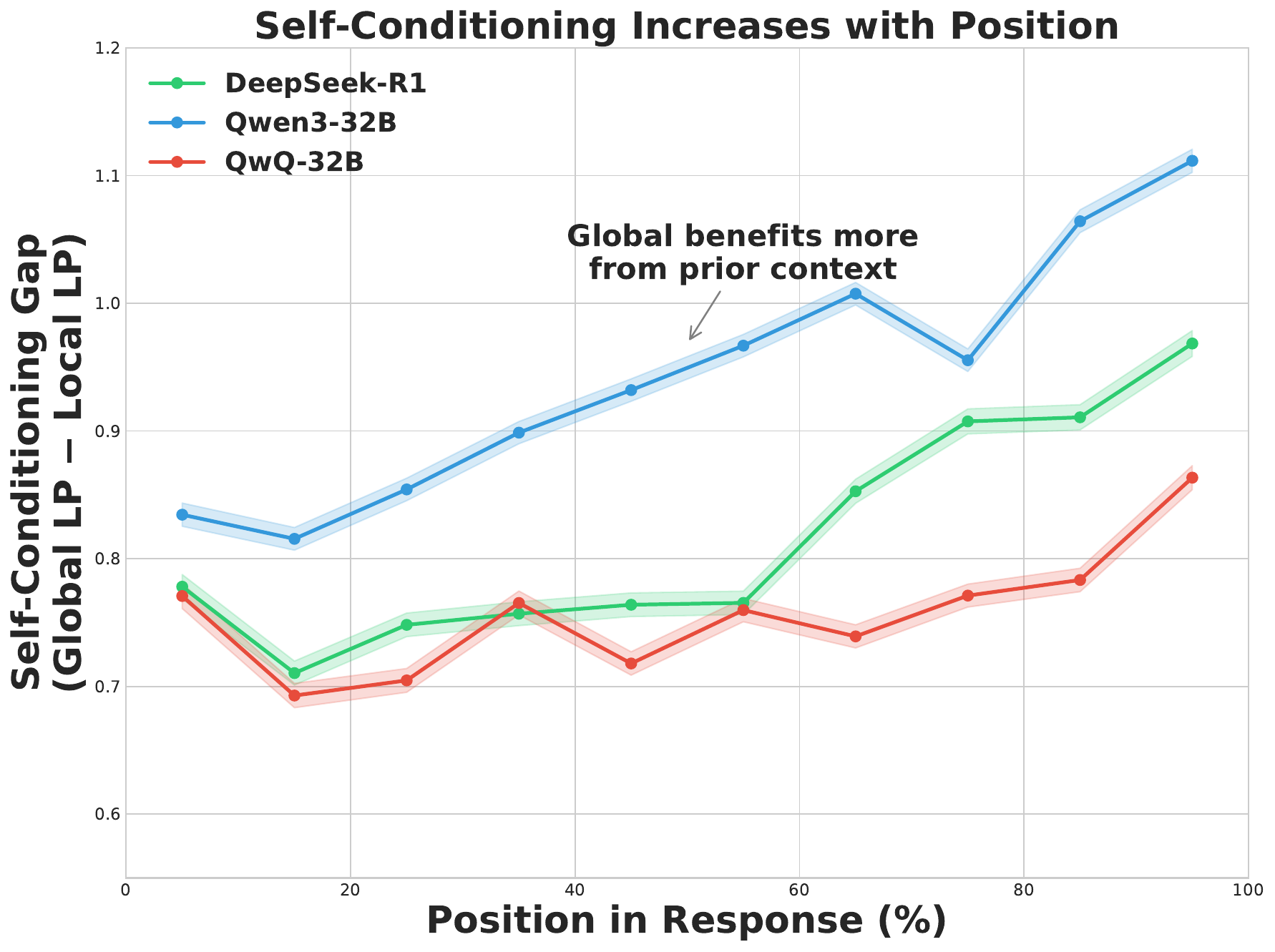}
    \caption{\textbf{Self-conditioning increases with position.} (a) The gap between global and local log probability grows monotonically across all three teachers, confirming that global scoring increasingly rewards tokens that echo earlier content. (b) Global LP rises with position (self-conditioning effect) while local LP remains relatively flat, demonstrating that LALP isolates reasoning difficulty from contextual repetition.}
    \label{fig:self_conditioning_gap}
\end{figure*}

Full-context scoring in GALP introduces a subtler problem: as context accumulates, later tokens become increasingly predictable simply by being consistent with earlier text—regardless of reasoning quality. Figure~\ref{fig:self_conditioning_gap} quantifies this self-reinforcement effect across all three teachers. Panel (a) shows that the gap between global and local log probability grows steadily with position: by the end of a response, global scoring assigns 0.8–1.1 nats more probability than local scoring, purely from accumulated context. Panel (b) reveals the underlying mechanism: global LP (dashed lines) rises throughout responses as the model becomes increasingly confident from self-referential context, while local LP (solid lines) remains flat, reflecting only the intrinsic difficulty of each step.
This self-conditioning explains why verbose, repetitive responses score highly under GALP: once the model commits to a pattern or intermediate value, restating it is nearly free. Qualitatively, we observe GALP-preferred responses frequently repeat earlier steps or values, and the model becomes increasingly confident each time (with examples below). LALP, by conditioning only on local context, scores each step for its intrinsic plausibility rather than its consistency with accumulated text.

\paragraph{Qualitative Evidence: Self-Conditioning in Action}
\label{subsec:qualitative_self_conditioning}

% \begin{figure}[h!]
%     \centering
%     \begin{minipage}{1.0\columnwidth}
%         \centering
%         \includegraphics[width=\textwidth]{figures/self_conditioning_examples.pdf}
%     \end{minipage}
%     \caption{Loss plots of the student model Qwen2.5-32B-Instruct trained on randomly selected data points from LIMO responses, highest local log probabilities responses, and highest global log probabilities responses.}
%     \label{fig:self condition examples}
% \end{figure}

Before presenting aggregate statistics, we illustrate self-conditioning through concrete examples from actual model generations. These examples demonstrate how global log probability inflates confidence for tokens that merely echo earlier content. We provide additional examples in Appendix due to space constraints.

\paragraph{Example 1: Numerical self-conditioning (``177'').}
Consider a response where the model calculates the final answer $m + n = 177$ three times. We track the probability of the last digit ``7'' across occurrences:

\begin{center}
\small
\begin{tabular}{lcc}
\toprule
\textbf{Occurrence} & \textbf{Log prob of ``7''} & \textbf{Probability} \\
\midrule
First calculation & $-1.83$ & 16.0\% \\
Verification step & $-0.10$ & 90.4\% \\
Final answer & $-0.00003$ & 99.99\% \\
\bottomrule
\end{tabular}
\end{center}

The probability jumps from 16\% to 99.99\% not because the model became more certain of the \emph{correctness} of the calculation, but because it is now copying from its own context. The initial 16\% probability reveals genuine computational uncertainty; the subsequent near-certainty reflects only self-conditioning.

\paragraph{Example 2: A toy demonstration.}
To isolate the phenomenon, consider the deliberately incorrect statement: ``1+1 is 3, so the answer is 3'':

\begin{center}
\small
\begin{tabular}{lcc}
\toprule
\textbf{Token} & \textbf{Log probability} & \textbf{Probability} \\
\midrule
``3'' (first occurrence) & $-3.01$ & 4.9\% \\
``3'' (second occurrence) & $-0.02$ & 97.7\% \\
\bottomrule
\end{tabular}
\end{center}

The model initially assigns low probability to the incorrect answer ``3'' (4.9\%), reflecting appropriate uncertainty. But after committing to this error, it assigns 97.7\% probability to repeating ``3'' as the final answer. Under GALP, this self-confirmed error receives a high average score, while LALP---by truncating context---can detect that the second ``3'' lacks local justification.

\paragraph{Implication.}
These examples reveal that global scoring conflates two distinct signals: (i) \emph{reasoning confidence} (whether a step is logically justified) and (ii) \emph{recall confidence} (whether a token matches earlier context). LALP separates these by evaluating each step with limited context, preventing prior commitments from inflating scores. We now quantify this effect systematically.

\paragraph{Quantitative Evidence: Self-Conditioning Gap vs Position}
\label{subsec:self_conditioning_gap}

The qualitative examples above suggest that self-conditioning should increase with position in the response: early tokens have little prior context to echo, while later tokens can leverage extensive earlier text. We test this prediction by measuring the \emph{self-conditioning gap}---the difference between global and local log probability---as a function of position across all responses.

\paragraph{Setup.}
For each token at position $t$ in a response, we compute:
\begin{equation}
\text{Gap}(t) = \log P_\text{global}(y_t | y_{<t}, x) - \log P_\text{local}(y_t | y_{t-k:t-1}, x)
\end{equation}
where $k$ is the local context window size. Positive gap indicates that global scoring assigns higher probability than local scoring---the signature of self-conditioning. We bin tokens by their relative position (0-100\% of response length) and compute mean gap per bin across ${\sim}800$ responses from each of three teacher models (DeepSeek-R1, QwQ-32B, Qwen3-32B).

\paragraph{Result.}
Figure~\ref{fig:self_conditioning_gap} confirms the prediction: the self-conditioning gap increases monotonically with position for all three teachers. At the beginning of responses (0-10\%), the gap is around 0.7-0.85. By the end (90-100\%), the gap increases by at least 0.1 to almost 0.3 across teachers, indicating that global scoring inflates probabilities for tokens late in responses.

\paragraph{Interpretation.}
The monotonic increase in self-conditioning gap has two important implications:
\begin{enumerate}[leftmargin=*,topsep=2pt,itemsep=1pt]
    \item \textbf{GALP overweights late tokens:} Later tokens contribute disproportionately high log probabilities to the global average, even when they merely repeat earlier content. This biases selection toward responses with more repetition in their latter portions.
    \item \textbf{LALP normalizes across position:} By truncating context, local scoring assigns similar probabilities to equivalent reasoning moves regardless of position, properly measuring step difficulty rather than recall ease.
\end{enumerate}

Qwen3 shows the largest self-conditioning gap at all positions, consistent with its verbose, repetitive style. This explains why Qwen3 achieves the highest GALP scores (Table~\ref{tab:teacher_selection_results}) despite producing the worst downstream student performance, its high global scores are inflated by self-conditioning rather than reflecting superior reasoning quality.

\subsubsection{Slower learning yet better generalization.}
\begin{figure*}[h!]
    \centering
    \includegraphics[width=0.89\linewidth]{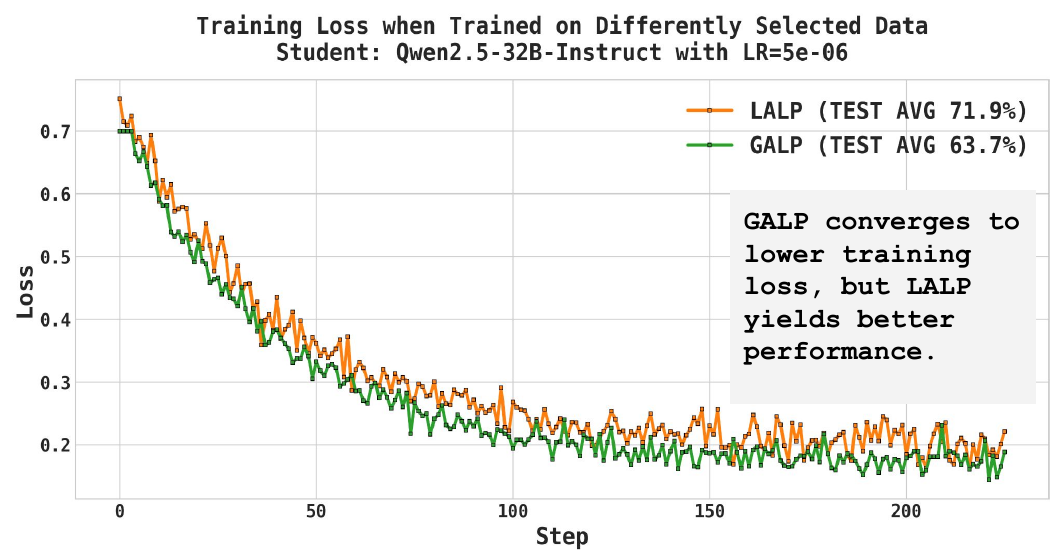}
    \caption{Training loss when fine-tuning on differently selected data (student: Qwen2.5-32B-Instruct). Global selection achieves lower training loss, but local selection yields higher downstream reasoning accuracy, illustrating a loss--generalization mismatch.}
    \label{fig:training_loss_paradox}
\end{figure*}

Figure~\ref{fig:training_loss_paradox} shows that GALP-selected data achieves lower training loss than LALP-selected data, yet yields worse downstream accuracy (Table~\ref{tab:data_selection_results_acrosss_teacher_models}). One interpretation is that the student fits globally fluent scaffolding easily, while the sparse reasoning transitions contribute minimally to the loss. Consistent with this view, LALP-selected data exhibits the opposite pattern: higher global training loss, but lower local NLL on held-out reasoning steps, suggesting the student has learned more reusable step-level patterns. 
% This dissociation supports the hypothesis that trajectory fluency is not the right target for reasoning—step-level learnability may matter more.
These findings support the hypothesis that reasoning performance is less dependent on trajectory fluency and more on the learnability of individual steps.

\section{Conclusion}

We identify a fundamental mismatch between how reasoning models generalize and how existing likelihood-based methods score candidate training data. While global average log probability (GALP) works well for short, single-teacher responses, it fails in the increasingly common regime of long reasoning traces from diverse teachers. The key issue is granularity: GALP scores entire trajectories, but students generalize by recombining familiar reasoning steps into novel solutions. Full-trajectory scoring rewards global fluency while the transferable signal lies in local step transitions. To fix this, we propose Local Average Log Probability (LALP). By scoring transitions within a small local window, LALP focuses on whether steps are justified by their immediate premises. This shift aligns data selection with how models actually learn, yielding up to a 9.4\% accuracy gain in math, better teacher selection, and strong cross-domain transfer to science and code.

\clearpage
\newpage

% % % % % \paragraph{Reproducibility statement.} The authors have made an effort to ensure the reproducibility of their work. The paper includes a detailed formulation of the method, along with all necessary hyperparameters for data generation, training, and evaluation. Furthermore, all codebases used in this research are properly referenced. In a commitment to fostering further research within the open-source community, the authors will also release the data and models.

\paragraph{Acknowledgement.} Ruoxi Jia and the ReDS lab acknowledge support through grants from the National Science Foundation under grants IIS-2312794, IIS-2313130, and OAC-2239622.

\section*{Impact Statement}
This paper advances the field of Machine Learning by exploring the intersection of data structuring and model generalization. While we do not identify specific immediate societal risks, the methodologies proposed, specifically logical segmentation and dynamic data selection, offer a path toward more interpretable multi-step inference systems. Future research may extend these principles beyond reasoning tasks to broader transfer learning settings, potentially improving how models use structured information in complex, real-world domains.
\textbf{Limitations.} LALP cannot verify long-range consistency, if an early step contains an error, later steps that correctly follow from it may still score highly. In settings where such consistency is critical, hybrid approaches combining local scoring with global coherence checks may be necessary.

\bibliography{iclr2026_conference}
\bibliographystyle{iclr2026_conference}

%%%%%%%%%%%%%%%%%%%%%%%%%%%%%%%%%%%%%%%%%%%%%%%%%%%%%%%%%%%%%%%%%%%%%%%%%%%%%%%
%%%%%%%%%%%%%%%%%%%%%%%%%%%%%%%%%%%%%%%%%%%%%%%%%%%%%%%%%%%%%%%%%%%%%%%%%%%%%%%
% APPENDIX
%%%%%%%%%%%%%%%%%%%%%%%%%%%%%%%%%%%%%%%%%%%%%%%%%%%%%%%%%%%%%%%%%%%%%%%%%%%%%%%
%%%%%%%%%%%%%%%%%%%%%%%%%%%%%%%%%%%%%%%%%%%%%%%%%%%%%%%%%%%%%%%%%%%%%%%%%%%%%%%
\newpage
\appendix
\onecolumn
\section{Toy theoretical model: locality induces a reasoning gap}
\label{app:toy_theory}

\paragraph{Motivation.}
This appendix provides a simple theoretical model supporting the main intuition of the paper: if training observations primarily expose \emph{local} dependencies between reasoning steps, then a learner trained with cross-entropy (plus smoothing/regularization) can learn local transitions but will collapse to an uninformative prior on long-range dependencies. This induces a ``reasoning gap'' that makes whole-trajectory scoring unreliable and motivates step-level scoring such as LALP.

\paragraph{Connection to empirical observations.}
Before presenting the formal model, we clarify how it connects to our main empirical findings. In Section~\ref{sec:analysis}, we observed that \emph{step-level embeddings} are densely covered by training data (cosine similarity 0.935), while \emph{trajectory-level embeddings} are sparsely covered (0.760). This coverage gap is the empirical manifestation of what we formalize below: contexts with substantial repeated support in training (local/step-like) allow the learner to acquire signal, while contexts that are effectively unseen (global/trajectory-like) force the learner to default to a prior.

The toy model below abstracts this as ``local observation,'' which should be understood as a proxy for \textbf{coverage/support density} rather than a literal claim that language models only see adjacent tokens. The key insight is that \emph{local reasoning patterns recur heavily across problems}, providing dense training signal, while \emph{specific trajectory configurations are rare}, leaving the learner underspecified at that granularity.

\paragraph{Setup (directed chain of steps).}
Let $Y_1,\dots,Y_N$ be discrete random variables taking values in a finite set $\mathcal{X}$ and assume the data distribution factorizes as a directed chain:
\begin{equation}
p_d(Y_1,\dots,Y_N)=p_d(Y_1)\prod_{j=1}^{N-1} p_d(Y_{j+1}\mid Y_j).
\label{eq:toy_chain_factorization}
\end{equation}
We interpret $Y_j$ as an \emph{atomic reasoning step}. The key assumption is \textbf{local observation}: training samples only expose local neighborhoods. Concretely, we observe pairs $(i,i+1)$ uniformly at random (or more generally, pairs within a window of size $k$), and we train a conditional model $q$ to predict a value given identifiers/previous observations.

\paragraph{Risk with smoothing.}
To model finite capacity, regularization, or inductive bias toward high-entropy predictors, consider the risk
\begin{equation}
R(q)=H(p,q)+\beta\,H(u,q),
\label{eq:toy_risk}
\end{equation}
where $H(\cdot,\cdot)$ is cross-entropy, $u$ is a uniform ``background'' distribution over the same support, and $\beta>0$ controls the strength of smoothing.\footnote{This form is equivalent to a cross-entropy term plus an entropy-regularization prior; it is also a standard way to model a learner that interpolates between empirical conditionals and a background distribution.}

\paragraph{Concrete distributions and model class.}
Let $\mathcal{I}=\{1,\dots,N\}$. We model a conditional predictor $q(\cdot \mid i,j,y)$ that outputs a distribution over $y'\in\mathcal{X}$ given an index pair $(i,j)\in\mathcal{I}^2$ and an observed value $y\in\mathcal{X}$ for $Y_i$. (This is the minimal abstraction needed for the locality argument.)

Let $p_{\text{obs}}$ be uniform over adjacent pairs: $p_{\text{obs}}(i,j)>0$ iff $j=i+1$. Define the \emph{local} data distribution $p$ over $(i,j,y,y')$ by
\[
p(i,j,y,y') = \begin{cases}
p_{\text{obs}}(i,j)\,p_d(Y_i=y,Y_{i+1}=y'), & j=i+1,\\
0, & \text{otherwise}.
\end{cases}
\]
Let $u$ be a uniform background distribution over the same variables (so in particular $u(y'\mid i,j,y)=1/|\mathcal{X}|$ for all contexts).

Finally, define cross-entropy for discrete distributions as $H(p,q)=\mathbb{E}_{X\sim p}[-\log q(X)]=-\sum_x p(x)\log q(x)$.

\begin{proposition}[Mixture minimizer]
\label{prop:mixture_minimizer}
Let $p_1$ and $p_2$ be discrete distributions and $\alpha_1,\alpha_2\ge 0$. The minimizer of $\alpha_1 H(p_1,q)+\alpha_2 H(p_2,q)$ over distributions $q$ is
\[
q^\star=\frac{\alpha_1}{\alpha_1+\alpha_2}p_1+\frac{\alpha_2}{\alpha_1+\alpha_2}p_2.
\]
\end{proposition}

\begin{proof}
Assume $p_1$ and $p_2$ have finite support $\Omega$ and $\alpha_1+\alpha_2>0$.\footnote{If $\alpha_1=\alpha_2=0$, then the objective is identically zero and any $q$ is a minimizer.}
Write the objective as
\[
J(q)=\alpha_1 H(p_1,q)+\alpha_2 H(p_2,q)
=-\sum_{x\in\Omega}\big(\alpha_1 p_1(x)+\alpha_2 p_2(x)\big)\log q(x),
\]
subject to $q(x)\ge 0$ and $\sum_x q(x)=1$.

Form the Lagrangian
\[
\mathcal{L}(q,\lambda)= -\sum_{x\in\Omega}\big(\alpha_1 p_1(x)+\alpha_2 p_2(x)\big)\log q(x)
\;+\;\lambda\Big(\sum_{x\in\Omega}q(x)-1\Big).
\]
For any $x$ with $q(x)>0$, the first-order condition is
\[
\frac{\partial \mathcal{L}}{\partial q(x)} = -\frac{\alpha_1 p_1(x)+\alpha_2 p_2(x)}{q(x)}+\lambda = 0,
\]
so
\[
q(x)=\frac{\alpha_1 p_1(x)+\alpha_2 p_2(x)}{\lambda}.
\]
Enforcing normalization yields
\[
1=\sum_{x\in\Omega} q(x)
 = \frac{1}{\lambda}\sum_{x\in\Omega}\big(\alpha_1 p_1(x)+\alpha_2 p_2(x)\big)
 = \frac{\alpha_1+\alpha_2}{\lambda},
\]
so $\lambda=\alpha_1+\alpha_2$. Substituting back gives
\[
q^\star(x)=\frac{\alpha_1 p_1(x)+\alpha_2 p_2(x)}{\alpha_1+\alpha_2},
\]
which is a valid distribution. Since $J(q)$ is strictly convex in $q$ on the probability simplex whenever $\alpha_1+\alpha_2>0$, this minimizer is unique.
\end{proof}

\begin{theorem}[Reasoning gap under locality]
\label{thm:reasoning_gap_locality}
Assume the observation process provides dense support only for adjacent pairs $(i,i+1)$ (or more generally, pairs within a window of size $k$), while non-local pairs have negligible support. Let $q^\star$ minimize \eqref{eq:toy_risk}. Then:
\begin{itemize}[leftmargin=*]
    \item For adjacent steps $(Y_i,Y_{i+1})$, the learned conditional interpolates between the true transition and the background:
    \[
    q^\star(\cdot\mid i,i+1,y)=\lambda_{i,y}\,p_d(\cdot\mid Y_i=y)+(1-\lambda_{i,y})\,u(\cdot)
    \quad\text{for some }\lambda_{i,y}\in(0,1).
    \]
    \item For non-adjacent steps $(Y_i,Y_j)$ with $|i-j|>1$, the model reverts to the background:
    \[
    q^\star(\cdot\mid i,j,y)=u(\cdot).
    \]
\end{itemize}
\end{theorem}

\paragraph{Interpretation.}
This result formalizes a general principle: \emph{where data provides dense support, the learner acquires signal; where support is sparse, the learner collapses to a prior}. In our empirical setting, ``local'' corresponds to step-level contexts that recur across problems (dense manifold coverage), while ``non-local'' corresponds to specific trajectory configurations that are rarely seen (sparse coverage). The theorem explains why GALP can be unreliable: it aggregates predictions over contexts where the model has no informative signal and defaults to a background distribution. LALP, by evaluating within local windows where support is dense, queries the model only where it has learned meaningful transitions.

\begin{proof}
We expand the risk \eqref{eq:toy_risk} by conditioning on contexts. Write $c=(i,j,y)$ for a context and let $p(c)$ denote the marginal probability of $c$ under $p$. Then
\begin{align*}
H(p,q)
&=\mathbb{E}_{(c,y')\sim p}\big[-\log q(y'\mid c)\big]\\
&=\sum_{c} p(c)\;\mathbb{E}_{y'\sim p(\cdot\mid c)}\big[-\log q(y'\mid c)\big]\\
&=\sum_{c} p(c)\;H\!\big(p(\cdot\mid c),\,q(\cdot\mid c)\big).
\end{align*}
Similarly,
\[
H(u,q)=\sum_{c} u(c)\;H\!\big(u(\cdot\mid c),\,q(\cdot\mid c)\big).
\]
Therefore,
\begin{equation}
R(q)=\sum_{c}\Big(p(c)\,H\!\big(p(\cdot\mid c),q(\cdot\mid c)\big)+\beta\,u(c)\,H\!\big(u(\cdot\mid c),q(\cdot\mid c)\big)\Big).
\label{eq:toy_risk_decomposition}
\end{equation}
Crucially, for each fixed context $c$, the distribution $q(\cdot\mid c)$ appears \emph{only} in the $c$-th summand. Hence, minimizing $R(q)$ over all conditionals $q(\cdot\mid c)$ reduces to minimizing each summand independently.

\paragraph{Case 1: non-adjacent pairs.}
If $j\ne i+1$, then by construction of $p$ we have $p(c)=0$. The $c$-th summand of \eqref{eq:toy_risk_decomposition} becomes $\beta\,u(c)\,H(u(\cdot\mid c),q(\cdot\mid c))$, which is minimized by $q(\cdot\mid c)=u(\cdot\mid c)=u(\cdot)$ (this also follows from Proposition~\ref{prop:mixture_minimizer} with $\alpha_1=0$, $\alpha_2=\beta u(c)$, and $p_2=u(\cdot\mid c)$).

\paragraph{Case 2: adjacent pairs.}
If $j=i+1$, then $p(c)=p_{\text{obs}}(i,i+1)\,p_d(Y_i=y)>0$ and
\[
p(\cdot\mid c)=p_d(\cdot\mid Y_i=y)
\]
by the chain factorization \eqref{eq:toy_chain_factorization}. The $c$-th summand of \eqref{eq:toy_risk_decomposition} is
\[
p(c)\,H\!\big(p_d(\cdot\mid Y_i=y),q(\cdot\mid c)\big)+\beta\,u(c)\,H\!\big(u(\cdot),q(\cdot\mid c)\big),
\]
which is minimized (by Proposition~\ref{prop:mixture_minimizer} with $\alpha_1=p(c)$, $\alpha_2=\beta u(c)$, $p_1=p_d(\cdot\mid Y_i=y)$, and $p_2=u(\cdot)$) by
\[
q^\star(\cdot\mid c)=\lambda_{i,y}\,p_d(\cdot\mid Y_i=y)+(1-\lambda_{i,y})\,u(\cdot),
\quad\text{where}\quad
\lambda_{i,y}=\frac{p(c)}{p(c)+\beta u(c)}\in(0,1).
\]
Combining both cases proves the theorem.
\end{proof}

\begin{theorem}[Signal dilution for GALP (fluency trap)]
\label{thm:signal_dilution_galp}
Fix a prompt $x$ and two candidate responses $y^{(c)}$ (``correct'') and $y^{(h)}$ (``hallucinated'') of equal length $m$ tokens. Let $\mathcal{I}_{\text{logic}}\subseteq\{1,\dots,m\}$ be indices of logic-bearing tokens with $|\mathcal{I}_{\text{logic}}|=N$ and let $\mathcal{I}_{\text{style}}$ be the complement.
Assume there exist constants $\delta>0$ and $\Delta>0$ such that, for all $t$,
\[
\log P_\theta\!\left(y^{(h)}_t \mid y^{(h)}_{<t}, x\right)-\log P_\theta\!\left(y^{(c)}_t \mid y^{(c)}_{<t}, x\right)\ge
\begin{cases}
\delta, & t\in \mathcal{I}_{\text{style}},\\
-\Delta, & t\in \mathcal{I}_{\text{logic}}.
\end{cases}
\]
Then the global average log-probability score (GALP) satisfies
\[
\mathrm{GALP}(y^{(h)}\mid x)>\mathrm{GALP}(y^{(c)}\mid x)
\quad\text{whenever}\quad
m>\frac{N(\Delta+\delta)}{\delta}.
\]
\end{theorem}

\begin{proof}
By definition (Equation~\ref{eqn:galp}), the score difference is
\[
\mathrm{GALP}(y^{(h)}\mid x)-\mathrm{GALP}(y^{(c)}\mid x)=\frac{1}{m}\sum_{t=1}^m\Big(\log P_\theta(y^{(h)}_t\mid y^{(h)}_{<t},x)-\log P_\theta(y^{(c)}_t\mid y^{(c)}_{<t},x)\Big).
\]
Split the sum into logic and style indices:
\[
\sum_{t=1}^m(\cdots)=\sum_{t\in\mathcal{I}_{\text{style}}}(\cdots)+\sum_{t\in\mathcal{I}_{\text{logic}}}(\cdots).
\]
Apply the assumed bounds term-by-term. For $t\in\mathcal{I}_{\text{style}}$, each summand is at least $\delta$, so the style sum is at least $(m-N)\delta$. For $t\in\mathcal{I}_{\text{logic}}$, each summand is at least $-\Delta$, so the logic sum is at least $-N\Delta$. Therefore,
\[
\mathrm{GALP}(y^{(h)}\mid x)-\mathrm{GALP}(y^{(c)}\mid x)\ge \frac{m-N}{m}\delta-\frac{N}{m}\Delta.
\]
The right-hand side is positive iff $(m-N)\delta>N\Delta$, i.e., $m>\frac{N(\Delta+\delta)}{\delta}$.
\end{proof}

\begin{theorem}[Step-level robustness of LALP]
\label{thm:lalp_robustness}
Let $y^{(c)}$ and $y^{(h)}$ be segmented into the same $p$ steps (sentences) $s_1,\dots,s_p$. Let $\ell_i(\cdot)$ denote the (token-average) log probability of step $s_i$ under the local window used by LALP (Equation~\ref{eq:local_log_prob}). Suppose there exists a ``fatal'' step index $i^\star$ and a gap $\gamma>0$ such that
\[
\ell_{i^\star}(y^{(c)})-\ell_{i^\star}(y^{(h)})\ge \gamma
\quad\text{and}\quad
\ell_i(y^{(c)})=\ell_i(y^{(h)}) \ \text{for all } i\neq i^\star.
\]
Then
\[
\mathrm{LALP}(y^{(c)}\mid x)-\mathrm{LALP}(y^{(h)}\mid x)\ge \frac{\gamma}{p}.
\]
\end{theorem}

\paragraph{Dilution comparison.}
Theorem~\ref{thm:signal_dilution_galp} shows that GALP dilutes by total tokens $m$, while Theorem~\ref{thm:lalp_robustness} shows that LALP dilutes by number of steps $p$. In our setting, $p \ll m$ (typically ${\sim}100$--$300$ steps vs.\ $10$K+ tokens), so LALP is \emph{far less diluted}. Note that if $p$ grows with $m$ (more sentences/steps in longer responses), the separation $\gamma/p$ does decay---but it decays with the number of \emph{reasoning moves}, not the number of \emph{tokens}, which is the key advantage.

\begin{proof}
By definition, $\mathrm{LALP}(y\mid x)=\frac{1}{p}\sum_{i=1}^p \ell_i(y)$. Hence,
\[
\mathrm{LALP}(y^{(c)}\mid x)-\mathrm{LALP}(y^{(h)}\mid x)
=\frac{1}{p}\sum_{i=1}^p\big(\ell_i(y^{(c)})-\ell_i(y^{(h)})\big).
\]
Under the assumptions, all terms cancel except the fatal step $i^\star$, so
\[
\mathrm{LALP}(y^{(c)}\mid x)-\mathrm{LALP}(y^{(h)}\mid x)=\frac{1}{p}\big(\ell_{i^\star}(y^{(c)})-\ell_{i^\star}(y^{(h)})\big)\ge \frac{\gamma}{p}.
\]
\end{proof}

\begin{corollary}[Robustness with bounded non-fatal differences]
\label{cor:lalp_robustness_eta}
In the setting of Theorem~\ref{thm:lalp_robustness}, replace the equality assumption for $i\neq i^\star$ with the weaker condition
\[
\ell_i(y^{(c)})-\ell_i(y^{(h)})\ge -\eta
\quad\text{for all } i\neq i^\star
\]
for some $\eta\ge 0$, while keeping $\ell_{i^\star}(y^{(c)})-\ell_{i^\star}(y^{(h)})\ge \gamma$. Then
\[
\mathrm{LALP}(y^{(c)}\mid x)-\mathrm{LALP}(y^{(h)}\mid x)\ge \frac{\gamma-(p-1)\eta}{p}.
\]
In particular, if $\gamma>(p-1)\eta$, then $\mathrm{LALP}(y^{(c)}\mid x)>\mathrm{LALP}(y^{(h)}\mid x)$.
\end{corollary}

\begin{proof}
From the identity in the proof of Theorem~\ref{thm:lalp_robustness},
\[
\mathrm{LALP}(y^{(c)}\mid x)-\mathrm{LALP}(y^{(h)}\mid x)
=\frac{1}{p}\sum_{i=1}^p\big(\ell_i(y^{(c)})-\ell_i(y^{(h)})\big).
\]
Lower bound the sum by separating the fatal step $i^\star$ and applying the assumptions:
\[
\sum_{i=1}^p(\cdots)
=\big(\ell_{i^\star}(y^{(c)})-\ell_{i^\star}(y^{(h)})\big)+\sum_{i\neq i^\star}\big(\ell_i(y^{(c)})-\ell_i(y^{(h)})\big)
\ge \gamma -(p-1)\eta.
\]
Dividing by $p$ yields the stated bound.
\end{proof}

\paragraph{Connection to GALP vs.\ LALP.}
This toy model highlights why whole-trajectory/global scoring can fail in long reasoning:
\begin{itemize}[leftmargin=*]
        \item \textbf{Global scoring queries under-supported contexts}: When the model has sparse coverage at the trajectory level, its predictions collapse toward a background distribution for those configurations, making global sequence scores less informative and easier to confound by fluent scaffolding (Theorem~\ref{thm:reasoning_gap_locality}).
    \item \textbf{Local scoring aligns with coverage}: LALP evaluates step transitions under a short window (Equation~\ref{eq:local_log_prob}), precisely where the model has dense support and nontrivial signal (the $\lambda p_d+(1-\lambda)u$ region). This is consistent with our empirical finding that teacher rankings are stable for small windows and converge to the pathological global ranking as the window grows (Figure~\ref{fig:context_window_vs_logprobs}).
    \item \textbf{Signal dilution compounds the problem}: Even when global scoring produces some signal, it is diluted by token-level averaging (Theorem~\ref{thm:signal_dilution_galp}), while LALP's step-level aggregation preserves sensitivity to reasoning-critical moves (Theorem~\ref{thm:lalp_robustness}).
\end{itemize}

\paragraph{Limitations of the toy model.}
We acknowledge several simplifications:
\begin{enumerate}[leftmargin=*]
    \item \textbf{Markov assumption}: The true data-generating process for reasoning may involve genuine long-range dependencies. Our model abstracts this as a chain to isolate the coverage effect; the key takeaway is about \emph{support density}, not the structure of true dependencies.
    \item \textbf{Uniform background}: In practice, the ``default'' when context is uninformative may not be uniform but could favor repetition/self-conditioning (the opposite of uniform). Our empirical analysis of self-reinforcement (Section~\ref{subsec:mechanisms}) addresses this complementary failure mode, which the theory does not capture.
    \item \textbf{Extension to window size $k$}: If the observation process exposes pairs within distance $k$, then the same argument implies a reasoning gap beyond distance $k$, matching our motivation for choosing a small local window for LALP.
\end{enumerate}

\section{Dataset and evaluation details}
\label{app:details}

\subsection{Step Segmentation Implementation}
\label{app:step_segmentation}

For decomposing responses into reasoning steps, we have tried using a deterministic sentence-based splitting heuristic. However, to improve correct step splitting with appropriate context, we adopt a capable, open-weight large language model, GLM-4.5-Air~\cite{Zeng2025GLM45AR}, which can handle large context that is necessary for our long reasoning data, to perform local step splitting with the following prompt:

\begin{verbatim}
Given the problem and solution to the problem, can you please split the 
solution into groups of logical sub steps and return those groups in a 
json format: 

    problem: {problem}
    solution: {solution}

    E.g. {
    "sentence_groups": {
        "group1": [
        "sentences from solution"
        ],
        "group2": [
        "sentences from solution"
        ],
        ...
    }
    }

Please split the solution into logical steps and return those steps in a json format.
Remember do not modify the solution phrasing, keep the wordings original and 
do not skip any step.
\end{verbatim}

% \paragraph{Splitting rules.}
% \begin{enumerate}
%     \item We split on standard sentence boundaries (periods, question marks, exclamation points) followed by whitespace.
%     \item LaTeX equation blocks delimited by `\$\$...\$\$` or `\textbackslash[...\textbackslash]` are treated as atomic units (not split even if they contain periods).
%     \item Numbered steps (e.g., ``1.'', ``Step 2:'') are preserved as split points.
%     \item Very short segments ($<$10 tokens) are merged with the preceding segment to avoid spurious splits from abbreviations (e.g., ``e.g.'', ``i.e.'').
% \end{enumerate}

% \paragraph{Implementation.} We use Python's NLTK sentence tokenizer with custom post-processing rules for the above LaTeX handling. For responses in plain text (non-LaTeX), standard sentence tokenization suffices.

% \paragraph{Sensitivity.} We conducted informal experiments with alternative segmentation strategies: (1) fixed 50-token chunks, (2) newline-based splits, and (3) manual annotation of 100 responses into semantic reasoning steps. Across these variants, LALP rankings showed Spearman correlation $> 0.85$, suggesting the metric is robust to reasonable changes in step granularity. However, a comprehensive sensitivity analysis across different step definitions remains important future work.

\subsection{Dataset Details}
\label{app:dataset}

\paragraph{Training Datasets.}  For training, we use two primary datasets. First, we use the MATH dataset \citep{hendrycks2021measuringmath}, and following prior works \citep{zeng2025simplerl,yu2025dapo}, we filter it to include only questions of difficulty levels 3-5, yielding 8,890 prompts (available at \url{https://huggingface.co/datasets/EleutherAI/hendrycks_math}). Second, to train models on reasoning data, we use the LIMO dataset \citep{ye2025limo}, a carefully curated collection of 817 prompts (available at \url{https://huggingface.co/datasets/GAIR/LIMO}).

\paragraph{Evaluation Datasets.}
To evaluate the performance of the model in mathematical capabilities, we include a wide suite of math benchmarks, including:
\begin{itemize}
    \item MATH500~\citep{hendrycks2021measuringmath} (500 Samples)\newline
    URL: \url{https://huggingface.co/datasets/EleutherAI/hendrycks_math}
    \item AIME 2025 (American Invitational Mathematics Examination) (30 Samples)\newline
    URL: \url{https://huggingface.co/datasets/opencompass/AIME2025}
    \item AMC 2023(American Mathematics Competition) (40 Samples)\newline
    URL: \url{https://huggingface.co/datasets/math-ai/amc23}
    \item MINERVA~\citep{lewkowycz2022solving} (272 Samples)\newline
    URL: \url{https://huggingface.co/datasets/knoveleng/Minerva-Math}
    \item KAOYAN (Chinese Graduate School Entrance Examinations) (199 Samples)\newline
    URL: \url{https://github.com/GAIR-NLP/LIMO/blob/main/eval/data/kaoyan/test.jsonl}
    \item OLYMPIADBENCH~\citep{he2024olympiadbench} (675 Samples)\newline
    URL: \url{https://huggingface.co/datasets/knoveleng/OlympiadBench}
    \item CN\_MATH\_2024 (Chinese High School Mathematics League Competition) (30 Samples)\newline
    URL: \url{https://github.com/GAIR-NLP/LIMO/blob/main/eval/data/cn_math_2024/test.jsonl}
    \item GPQA-D (A Graduate-Level Google-Proof Q\&A Benchmark) (198 Samples) \newline
    URL: \url{https://huggingface.co/datasets/Idavidrein/gpqa}
    \item LCBv2 (LiveCodeBench) (511 Samples) \newline
    URL: \url{https://github.com/LiveCodeBench/LiveCodeBench}
\end{itemize}

% \paragraph{Problem statement.} Given a set of candidate responses $Y = \{y^{(1)}, y^{(2)}, \dots, y^{(k)}\}$ generated from diverse sources for a specific input prompt $x$, our goal is to select the most suitable response $y^* \in Y$ for SFT of a student $S$ using the pair $(x,y^*)$. In this paper, we primarily focus on the mathematical reasoning domain. This domain provides access to long, structured reasoning data and allows for straightforward verification of response correctness. 

\subsection{Model Details}
\label{app:models}

\paragraph{Student Models.} For student models, we perform supervised fine-tuning on:
\begin{itemize}
    \item Qwen2.5-Math-7B \citep{yang2024qwen25mathtechnicalreportmathematical}\newline
    URL: \url{https://huggingface.co/Qwen/Qwen2.5-Math-7B}
    \item Qwen2.5-7B-Instruct \citep{qwen2, yang2024qwen2} \newline
    URL: \url{https://huggingface.co/Qwen/Qwen2.5-7B-Instruct}
    \item Qwen2.5-32B-Instruct \citep{qwen2, yang2024qwen2} \newline
    \item Llama-3.1-8B-Instruct \citep{grattafiori2024llama} \newline
    URL: \url{https://huggingface.co/meta-llama/Llama-3.1-8B-Instruct}
\end{itemize}

\paragraph{Teacher Models.} For teacher models, we sample responses from the following models:
\begin{itemize}
    \item Qwen2.5-72B-Instruct \citep{qwen2, yang2024qwen2} \newline
    URL: \url{https://huggingface.co/Qwen/Qwen2.5-32B-Instruct}
    \item Gemma3-27B-IT \citep{team2025gemma} \newline
    URL: \url{https://huggingface.co/google/gemma-3-27b-it}
    \item DeepSeek-R1 \citep{DeepSeekAI2025DeepSeekR1IR} \newline
    URL: \url{https://huggingface.co/deepseek-ai/DeepSeek-R1}
    \item QWQ-32B\citep{qwq32b} \newline
    URL: \url{https://huggingface.co/Qwen/QWQ32b}
    \item Qwen3-32B \citep{yang2025qwen3technicalreport} \newline
    URL: \url{https://huggingface.co/Qwen/Qwen3-32B}
\end{itemize}

\subsection{Experimental Details} 
\label{app:exp_detail}

\paragraph{Sampling Hyperparameters.} Training data for fine-tuning student models were generated by sampling outputs from teacher models. We use the vLLM library~\citep{kwon2023efficient} for this process to ensure efficient inference, employing the sampling hyperparameters detailed in Table~\ref{tab:sampling hyper}.

\begin{table}[h!]
    \centering
    \begin{tabular}{lr}
        \toprule
        Property & Value\\ \midrule
         Number of samples & 1/16 \\
         Temperature & 0.0/1.0 \\
         Top P & 1.0/0.95 \\
         Top K & 1/40 \\ 
         Max Tokens & 42786+ \\ 
         \bottomrule \\
    \end{tabular}
    \caption{The hyperparameters for sampling from the teacher models using vLLM~\citep{kwon2023efficient}. }
    \label{tab:sampling hyper}
\end{table}

% vllm

% sampling_params = SamplingParams(n=n_samples, temperature=0.6, 
%                          top_p=0.95, top_k=40,
%                              max_tokens=82500)

\paragraph{Training Hyperparameters.} For supervised fine-tuning on student models, we leverage the LLaMA-Factory~\citep{zheng2024llamafactory} platform that offers efficient training and apply the following setting of hyperparameters (listed in Table~\ref{tab:training hyper}):

\begin{table}[h!]
    \centering
    \begin{tabular}{lr}
        \toprule
        Property & Value\\ \midrule
         Train Batch Size Per Device & 1/2 \\
         Gradient Accumulation Steps & 8 \\
         Learning Rate & $5.0\times10^{-6}$/$1.0\times10^{-5}$ \\
         Epochs & 10/15 \\ 
         Warmup Ratio & 0.0 \\ 
         BFloat16  & True \\ 
         \bottomrule \\
    \end{tabular}
    \caption{The hyperparameters for SFT the student models using LLaMA Factory~\citep{zheng2024llamafactory}. }
    \label{tab:training hyper}
\end{table}

% https://github.com/hiyouga/LLaMA-Factory

\paragraph{Evaluation Hyperparameters.} After models are trained, we evaluate the models on a variety of mathemtical benchmarks using the evaluation library from LIMO~\citep{ye2025limo} (URL: \url{https://github.com/GAIR-NLP/LIMO/tree/main/eval}) with the following hyperparameters (Table~\ref{tab:evaluation hyper}):

After training, we evaluate the models on a range of mathematical benchmarks using the evaluation library provided by LIMO~\citep{ye2025limo} based on the Qwen2.5-Math evaluation code~\citep{yang2024qwen2} (available at \url{https://github.com/GAIR-NLP/LIMO/tree/main/eval}). The evaluation is conducted using the hyperparameter settings from DeepSeek-R1
\cite{DeepSeekAI2025DeepSeekR1IR} as detailed in Table~\ref{tab:evaluation hyper}. 

\begin{table}[h!]
    \centering
    \begin{tabular}{lr}
        \toprule
        Property & Value\\ \midrule
         Temperature & 0.0/0.6 \\
         Max Tokens & 32768 \\
         Top P & 1/0.95 \\
         Pass@K & 1/8 \\
         Samples & 1/8 \\
         \bottomrule \\
    \end{tabular}
    \caption{The hyperparameters for evaluation of the student models at the inference stage using evaluation code from Qwen2.5-Math evaluation~\citep{yang2024qwen2}.}
    \label{tab:evaluation hyper}
\end{table}

% https://github.com/GAIR-NLP/LIMO/tree/main/eval

\paragraph{Software and Hardware.} In our experiments we used NVIDIA 4xA100 GPUs for training and evaluation. For reproducibility of our results, we share our code in an anonymized repository for the submission purposes: \hyperlink{https://anonymous.4open.science/r/lalp-5272/}{https://anonymous.4open.science/r/lalp-5272/}.

\section{Additional Results}
\label{app: results}

\subsection{Loss Comparison}
\label{app:loss_comparison}

We compare the loss curves of student models trained on data selected using global and local log likelihood criteria, as summarized in Table~\ref{tab:data_selection_results_acrosss_teacher_models}. The corresponding loss plots are presented in Figure~\ref{fig:training_loss_paradox}.

% \begin{figure}[h!]
%     \centering
%     \begin{minipage}{1.0\textwidth}
%         \centering
%         \includegraphics[width=\textwidth]{figures/loss_qwen32b.png}
%     \end{minipage}
%     \caption{Loss plots of the student model Qwen2.5-32B-Instruct trained on randomly selected data points from LIMO responses, highest local log probabilities responses, and highest global log probabilities responses.}
%     \label{fig:loss landscapes}
% \end{figure}

As shown in Figure~\ref{fig:training_loss_paradox}, models trained on responses with the highest global log-likelihood demonstrate the fastest convergence and lowest training loss compared to those trained on randomly selected data or responses with the highest local log-likelihood. This behavior is expected, as high global log-likelihood responses likely represent more cohesive and natural samples as a whole, which align more closely with the student model's existing representation space. Such data may provide clearer learning signals, enabling the model to fit the training distribution more efficiently. However, as shown earlier in Table~\ref{tab:data_selection_results_acrosss_teacher_models}, instead, the model trained on data selected by highest local log-likelihood ultimately achieves better downstream performance. This highlights a key insight: while global log-likelihood data may facilitate faster convergence during training, this does not necessarily translate to better generalization, underscoring the limitations of relying solely on loss curves as indicators of final model performance.

\subsection{Data Composition From Selection}
\label{app:data_composition}
In Section~\ref{sec:experiments}, we have chosen LIMO responses across three teachers (DeepSeek-R1, QWQ-32B, Qwen3-32B) based on local and global naturalness and provided results in Table~\ref{tab:data_selection_results_acrosss_teacher_models}. Here, we provide the composition of selected responses across teachers depending on the method in Table~\ref{tab: data composition}.

\begin{table}[h!]
    \centering
    \begin{tabular}{lccc}
    \toprule
                        & DeepSeek-R1 & QWQ-32B& Qwen3.0-32B \\
                        \midrule
                & \multicolumn{3}{c}{{\textbf{Student: Qwen2.5-32B-Instruct}}}   \\
                \midrule
         Random &  33.3 & 33.3 & 33.4\\
         Local Lowest &  42.4 & 11.3 & 46.3\\
         Global Highest &  47.6 & 7.2 & 45.2\\
         Local Highest &  42.4 & 36.3 & 21.3  \\
                        \midrule
                & \multicolumn{3}{c}{{\textbf{Student: Qwen2.5-32B-Instruct}}}   \\
                        \midrule
         Random &  33.3 & 33.3 & 33.4\\
         Local Lowest &  43.3 & 20.4 & 36.3\\
         Global Highest &  47.2 & 8.6 & 44.2\\
         Local Highest &  26.8 & 44.9 & 28.3\\
         \bottomrule
    \end{tabular}
    \caption{Data composition from different teacher models for the LIMO responses depending on the selection method(\%).}
    \label{tab: data composition}
\end{table}

\subsection{Additional Examples on Self-Conditioning}
\label{app:self condition examples}

\paragraph{Example 3: Multi-digit number recall (``1228'').}
A more dramatic example: the model computes $(2457 - 1)/2 = 1228$ and subsequently references this value:

\begin{center}
\small
\begin{tabular}{lcc}
\toprule
\textbf{Context} & \textbf{Log prob of ``1''} & \textbf{Probability} \\
\midrule
First calculation: ``$= 1228$'' & $-0.33$ & 71\% \\
Recall: ``is 1228 divisors'' & $-0.00006$ & 99.99\% \\
Usage: ``1228 - 639 = ...'' & $-0.006$ & 99.4\% \\
\bottomrule
\end{tabular}
\end{center}

The model transitions from \emph{calculating} (71\% confidence) to \emph{copying} (99.99\% confidence). Under global scoring, all three occurrences contribute to the average log probability, but only the first reflects genuine reasoning difficulty.

\paragraph{Example 4: Stylistic self-conditioning (``come in pairs'').}
Self-conditioning affects not only numbers but also phrasing choices:

\begin{center}
\small
\begin{tabular}{lcc}
\toprule
\textbf{Context} & \textbf{Log prob of ``come''} & \textbf{Probability} \\
\midrule
First use: ``divisors ... come in pairs'' & $-1.28$ & 27\% \\
Second use: ``divisors come in pairs'' & $-0.66$ & 51\% \\
\bottomrule
\end{tabular}
\end{center}

The model nearly doubles its confidence in the word ``come'' simply because it used that phrasing earlier. This stylistic self-conditioning inflates GALP scores for responses with repetitive phrasing patterns.

\subsection{Breakdown of results in main paper.}

\begin{table}[h!]
\resizebox{\columnwidth}{!}{%
\begin{tabular}{lrrrrrrrrrr}
\textbf{}        & \multicolumn{1}{l}{\textbf{MATH}} & \multicolumn{1}{l}{\textbf{AIME25}} & \multicolumn{1}{l}{\textbf{AMC}} & \multicolumn{1}{l}{\textbf{MINERVA}} & \multicolumn{1}{l}{\textbf{KAOYAN}} & \multicolumn{1}{l}{\textbf{OLYMPIADB}} & \multicolumn{1}{l}{\textbf{CN\_MATH24}} & \multicolumn{1}{l}{\textbf{AVG}} & \textbf{GALP }                 & \textbf{LALP}                   \\
\toprule
\multicolumn{9}{c}{{\textbf{Student: Qwen2.5-7B-Instruct}}}                                                                                                                                                                                                                                                                                 &                            &                            \\
\toprule
\textbf{Student Before SFT}  & 0.752                             & 0.167                               & 0.500                            & 0.268                                 & 0.216                               & 0.404                                      & 0.167                                       & 0.353                            & \multicolumn{1}{c}{-}                          & \multicolumn{1}{c}{-}                          \\
\midrule
\textbf{Qwen3-32B Data}    & 0.714                             & 0.166                               & 0.500                            & 0.279                                 & 0.389                               & 0.375                                      & 0.133                                       &  \cellcolor{red!20}0.365                            &   \cellcolor{green!20}  -0.697                     & \cellcolor{red!20} -0.279                     \\
\textbf{DeepSeek-R1 Data}  & 0.784                             & 0.166                               & 0.600                            & 0.239                                 & 0.330                               & 0.441                                      & 0.233                                       & \cellcolor{yellow!20} 0.399                            & \cellcolor{red!20} -0.796                     &\cellcolor{yellow!20} -0.264                     \\
\textbf{QWQ-32BData}       & 0.780                             & 0.266                               & 0.600                            & 0.275                                 & 0.356                               & 0.442                                      & 0.2                                         & \cellcolor{green!20}0.417                            & \cellcolor{yellow!20}  -0.743                     & \cellcolor{green!20} -0.241                     \\
\toprule
\multicolumn{9}{c}{{\textbf{Student: Qwen2.5-32B-Instruct}}}                                                                                                                                                                                                                                                                       &                            &                            \\
\toprule
\textbf{Student Before SFT}  & 0.822                             & 0.133                               & 0.700                            &          0.298                             &            0.422                         &                 0.471                           &            0.233                                 &          0.445                        & \multicolumn{1}{c}{-}      & \multicolumn{1}{c}{-}      \\
\midrule
\textbf{Qwen3-32B Data}    & 0.882                             & 0.567                               & 0.900                            & 0.353                                 & 0.598                               & 0.559                                      & 0.600                                       & \cellcolor{red!20}0.637                            & \cellcolor{green!20}-0.800 & \cellcolor{red!20}-0.257 \\
\textbf{DeepSeek-R1 Data}  & 0.896                             & 0.467                               & 0.925                            & 0.338                                 & 0.613                               & 0.644                                      & 0.733                                       & \cellcolor{yellow!20}0.659                            & \cellcolor{red!20}-0.895 & \cellcolor{yellow!20}-0.241 \\
\textbf{QWQ-32BData}       & 0.916                             & 0.633                               & 0.975                            & 0.364                                 & 0.653                               & 0.689                                      & 0.800                                       & \cellcolor{green!20}0.719                            & \cellcolor{yellow!20}-0.888 & \cellcolor{green!20}-0.218  
                                       \\ 
                                       \bottomrule
\end{tabular}
}
\caption{Teacher selection results. LALP correctly ranks teachers by downstream performance (QwQ > DeepSeek-R1 > Qwen3), while GALP produces an inverted ranking. Colors: \colorbox{green!20}{best}, \colorbox{yellow!20}{middle}, \colorbox{red!20}{worst}.}\label{tab:teacher_selection_results_full}
\vspace{-1.5em}
\end{table}

\begin{table}[h!]
\centering
\resizebox{0.90\columnwidth}{!}{%
\begin{tabular}{lllllllll}
\textbf{}                & \multicolumn{1}{c}{\textbf{MATH}} & \multicolumn{1}{c}{\textbf{AIME25}} & \multicolumn{1}{c}{\textbf{AMC}} & \multicolumn{1}{c}{\textbf{MINERVA}} & \multicolumn{1}{c}{\textbf{KAOYAN}} & \multicolumn{1}{c}{\textbf{OLYMPIAD}} & \multicolumn{1}{c}{\textbf{CN\_MATH24}} & \multicolumn{1}{c}{\textbf{AVG}} \\
\toprule
\multicolumn{9}{c}{{\textbf{Student: Qwen2.5-7B-Instruct}}}                                                                                                                                                                                                                                                                                         \\
\toprule
\textbf{Original Model} & 0.752       & 0.167         &  0.500       & 0.268           & 0.216     & 0.404                  & 0.167               & 0.353      \\
\midrule
\textbf{Random}          & 0.768                             & 0.133                               & 0.625                            & 0.268                                 & 0.367                               & 0.456                                      & 0.233                                       & 0.407                            \\
\textbf{GALP}  & 0.762                             & 0.2                                 & 0.6                              & 0.268                                 & 0.381                               & 0.441                                      & 0.233                                       & 0.412                            \\
\textbf{Local Lowest}    & 0.742                             & 0.167                               & 0.575                            & 0.298                                 & 0.342                               & 0.433                                      & 0.233                                       & 0.399                            \\
\textbf{LALP}   & 0.788                             & 0.2                                 & 0.625                            & 0.298                                 & 0.392                               & 0.441                                      & 0.333                                       & \textbf{0.440}                            \\
\toprule
\multicolumn{9}{c}{{\textbf{Student: Qwen2.5-32B-Instruct}}}                                                                                                                                                                                                                                                                               \\
\toprule
\textbf{Original Model} & 0.824        & 0.133          & 0.700       & 0.298             & 0.422        & 0.471          & 0.233          & 0.445      \\
\midrule
\textbf{Random}          &   0.906      &0.400         &     0.925   &     0.327    &        0.628        &     0.636     &       0.733     &     0.651   \\
\textbf{GALP}  &    0.876     &  0.433       &   0.825     &   0.331       &       0.592        &   0.636       &      0.733      &     0.632    \\
\textbf{Local Lowest}    &    0.896     &   0.400      &    0.825    &     0.324     &      0.608         &    0.640      &       0.700       &     0.623 \\
\textbf{LALP}   &   0.902     &   0.667     &     1.000    &     0.353       &     0.653         &   0.673      &      0.833       &   \textbf{0.726}
                                       \\ 
                                       \bottomrule       
\end{tabular}
}
\caption{Response selection from mixed-teacher pool. LALP (Local Highest) outperforms GALP (Global Highest) by \textbf{+9.4\%} on the 32B student (0.726 vs 0.632). Notably, LALP-selected data even outperforms training on all responses from the best single teacher (QwQ-32B: 0.719 in Table~\ref{tab:teacher_selection_results}).}
\label{tab:data_selection_results_acrosss_teacher_models_full}
% \vspace{-1em}
\end{table}

\subsection{Additional Results on MATH prompts}
\label{app:math_prompts}
We present additional results for Qwen2.5-7B-Instruct on the MATH benchmark, using responses generated by two different teacher models: Qwen2.5-72B-Instruct, which tends to produce shorter responses, and QWQ-32B, which generates longer reasoning responses. We provide a comprehensive summary of these results in Table~\ref{tab:within_teacher_non_reasoning_results}.

% \subsection{Ablation: Context Window Size and Teacher Ranking}
% \label{app:context_window_ranking}

% A key design choice in LALP is the context window size $k$. We analyze how the teacher ranking changes as $k$ increases from local to global.

% \paragraph{Setup.} We compute local log probabilities for LIMO responses from all three teachers using Qwen-7B-Instruct as the student, varying the context window from 5\% to 75\% of total sentences.

% \begin{figure}[h!]
%     \centering
%     \includegraphics[width=0.65\textwidth]{figures/relatinoship.png}
%     \caption{Teacher ranking by log probability as context window grows. At small windows (5--25\%), LALP correctly ranks QWQ highest (best downstream performance). As windows approach global (50--75\%), the ranking inverts to match GALP, recovering the pathological ordering where Qwen3-32B (worst performance) ranks first.}
%     \label{fig:context_window_vs_logprobs_app}
% \end{figure}

% \paragraph{Key finding.} As shown in Figure~\ref{fig:context_window_vs_logprobs_app}, small context windows (5--25\%) produce a stable ranking that correctly identifies QWQ as the best teacher. As $k$ increases toward 50--75\%, the rankings converge to the global ordering---recovering the pathological case where Qwen3-32B (highest global score, worst performance) ranks first. This proves that \emph{locality itself} is the mechanism through which LALP avoids the fluency trap.

\subsection{Ablation: Test Set Coverage}
\label{app:test set coverage}
\begin{figure}[h!]
    \centering
    \begin{minipage}{0.8\textwidth}
        \centering
        \includegraphics[width=\textwidth]{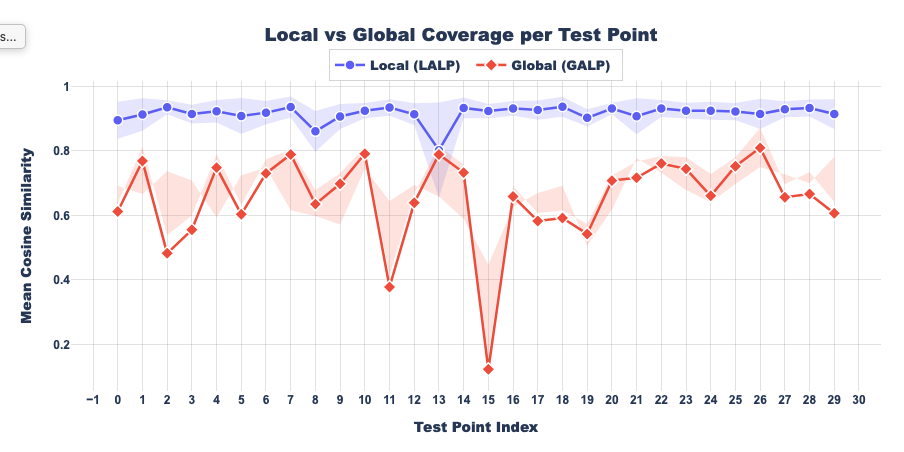}
    \end{minipage}
    \caption{AMC Test Points}
    \label{fig:contextamc}
\end{figure}

\begin{figure}[h!]
    \centering
    \begin{minipage}{0.8\textwidth}
        \centering
        \includegraphics[width=\textwidth]{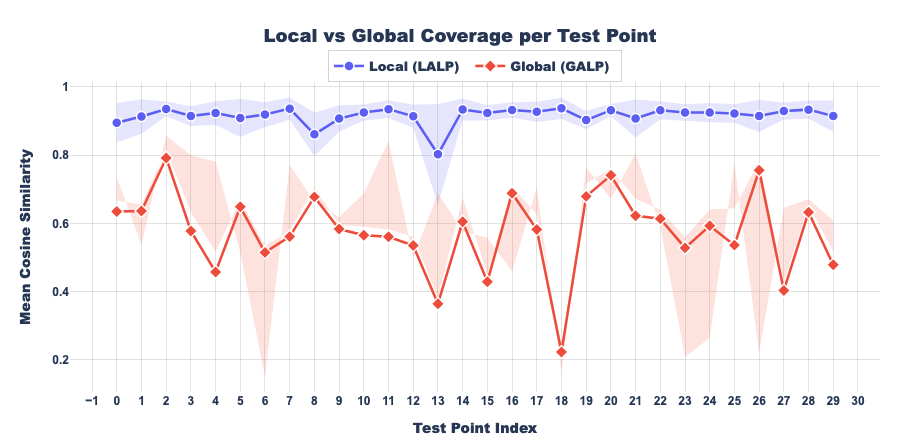}
    \end{minipage}
    \caption{GSM8K Test Points}
    \label{fig:contextgsm}
\end{figure}

Here we present other test sets that also show the gap between GALP and LALP behavior in Figures \ref{fig:contextamc} and \ref{fig:contextgsm}.

\subsection{Ablation: Context Window Size vs Performance}
\label{app:context_window_size_vs_performance}
\begin{figure}[h!]
    \centering
    \begin{minipage}{0.8\textwidth}
        \centering
        \includegraphics[width=\textwidth]{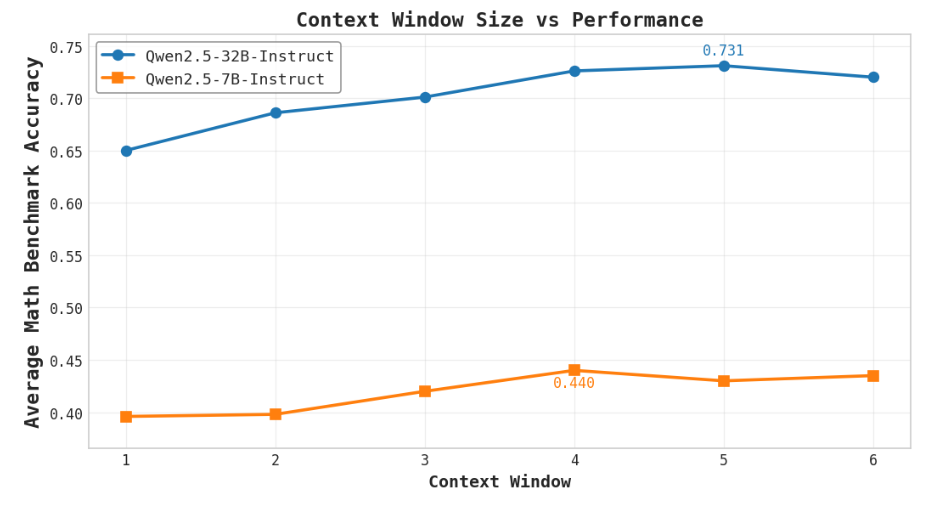}
    \end{minipage}
    \caption{Context window size vs performance. Ablation on two student models.}
    \label{fig:context}
\end{figure}

We also provide an ablation study of the context window size in terms of downstream performance. As we observe in Figure~\ref{fig:context}, there exists an optimal window size that balances local context with stylistic independence.

\subsection{Qualitative Examples: Global vs Local Nearest Neighbors}
\label{app:qualitative_examples}

To provide intuition for the quantitative coverage results, we present qualitative nearest-neighbor examples comparing trajectory-level (global) and step-level (local) embeddings from AIME 2025 test problems.

\paragraph{Global (Trajectory-Level) Examples.}
When matching \emph{full solution trajectories}, the nearest training neighbors often differ substantially in problem structure and reasoning approach. This reflects the sparse manifold coverage (cosine similarity 0.760) reported in Figure~\ref{fig:coverage_global_vs_local}.

\begin{table}[h!]
\centering
\small
\begin{tabular}{p{0.45\textwidth}|p{0.45\textwidth}}
\toprule
\textbf{AIME Test Trajectory (truncated)} & \textbf{Nearest Training Trajectory (truncated)} \\
\midrule
``Let me work through this combinatorics problem systematically. We need to count the number of ways to arrange...'' (cos=0.72) & ``I'll solve this number theory problem step by step. First, let's identify the prime factorization...'' \\
\midrule
``This is a geometry problem involving circles and tangent lines. Let me set up coordinates...'' (cos=0.68) & ``For this probability question, I need to compute the expected value by considering all possible outcomes...'' \\
\bottomrule
\end{tabular}
\caption{Global nearest-neighbor examples: full trajectories match poorly across problem types.}
\label{tab:global_nn_examples}
\end{table}

\paragraph{Local (Step-Level) Examples.}
In contrast, when matching \emph{individual reasoning steps}, nearly every test step finds a highly similar training step. This reflects the dense manifold coverage (cosine similarity 0.935) that enables compositional generalization.

\begin{table}[h!]
\centering
\small
\begin{tabular}{p{0.45\textwidth}|p{0.45\textwidth}}
\toprule
\textbf{AIME Test Step} & \textbf{Nearest Training Step} \\
\midrule
``Since $\gcd(a,b)=1$, we can apply B\'ezout's identity to find integers $x,y$ such that $ax+by=1$.'' (cos=0.96) & ``Because $\gcd(m,n)=1$, B\'ezout's lemma guarantees integers $u,v$ with $mu+nv=1$.'' \\
\midrule
``Substituting $x=2$ into the equation gives $f(2)=4+3(2)-7=3$.'' (cos=0.94) & ``Plugging in $t=2$ yields $g(2)=2^2+5(2)-4=10$.'' \\
\midrule
``By the triangle inequality, $|a+b| \leq |a|+|b|$.'' (cos=0.98) & ``The triangle inequality gives us $|x+y| \leq |x|+|y|$.'' \\
\midrule
``Let $S_n$ denote the sum of the first $n$ terms.'' (cos=0.97) & ``Define $T_k$ as the sum of the first $k$ elements.'' \\
\bottomrule
\end{tabular}
\caption{Local nearest-neighbor examples: individual reasoning steps match closely across problems, enabling compositional reuse.}
\label{tab:local_nn_examples}
\end{table}

\paragraph{Interpretation.}
These examples illustrate why step-level data selection outperforms trajectory-level selection. While complete solutions to novel AIME problems are genuinely new (no close training match exists), the \emph{atomic reasoning moves}---algebraic substitutions, theorem applications, definitional setups---are well-represented in training data. LALP leverages this compositional structure by scoring responses based on how well-supported each individual step is, rather than requiring a global template match that rarely exists.

\subsection{GALP Works Within Single Teacher (Short Responses)}
\label{app:galp_single_teacher}

We replicate the finding from prior work~\citep{zhang2025best} that global log probability selection works well within a single teacher on shorter responses. This establishes the baseline validity of GALP before examining its failure modes.

\paragraph{Setup.} We use MATH prompts of level 3-5 difficulty (8,890 prompts). For each prompt, we generate 16 responses using a teacher model with temperature 0.6 and top-p 0.95. We select responses with highest, middle, and lowest global log probability for comparison. Students are fine-tuned for 5 epochs with learning rate 1e-5.

\begin{table*}[h!]
\resizebox{\textwidth}{!}{%
\begin{tabular}{llrrrrrrrr}
                                       &   & \textbf{MATH}             & \multicolumn{1}{l}{\textbf{AIME25}} & \multicolumn{1}{l}{\textbf{AMC}} & \multicolumn{1}{l}{\textbf{MINERVA}} & \multicolumn{1}{l}{\textbf{KAOYAN}} & \multicolumn{1}{l}{\textbf{OLYMPIADB}} & \multicolumn{1}{l}{\textbf{CN\_MATH24}} & \multicolumn{1}{l}{\textbf{AVG}}  \\ 
                                       \toprule
                                       & \multicolumn{9}{c}{\textbf{Student: Qwen2.5-7B-Instruct}}                                                                                                                                                                                                                                                                             \\ \toprule
                                       & Original Model           & 0.752                               & 0.167                            & 0.5                                   & 0.268                               & 0.216                                      & 0.404                                       & 0.167                            & 0.353          \\
                                       \midrule
\textbf{Teacher:} & Lowest LP  & 0.678                               & 0.1                              & 0.3                                   & 0.224                               & 0.296                                      & 0.314                                       & 0.133                            & 0.292                \\
\textbf{Qwen2.5-72B}                                       & Middle LP  & 0.71                                & 0.1                              & 0.425                                 & 0.257                               & 0.336                                      & 0.339                                       & 0.2                              & 0.338                \\
\textbf{-Instruct}  
                                       & Highest LP & 0.744                               & 0.133                            & 0.5                                   & 0.252                               & 0.391                                      & 0.391                                       & 0.167                            & 0.368                \\
                                       \midrule
\textbf{Teacher: }        & Lowest LP  & 0.667                               & 0.1                              & 0.375                                 & 0.165                               & 0.226                                      & 0.29                                        & 0.133                            & 0.279           \\
\textbf{Gemma3-27B}                                        & Middle LP & 0.716                               & 0.1                              & 0.475                                 & 0.129                               & 0.246                                      & 0.357                                       & 0.167                            & 0.313        \\
\textbf{-IT}
                                       & Highest  LP & 0.712                               & 0.1                              & 0.5                                   & 0.176                               & 0.251                                      & 0.362                                       & 0.133                            & 0.319          \\
                                       \toprule
                                       & \multicolumn{9}{c}{\textbf{Student: Qwen2.5-Math-7B}}                                                                                                                                                                                                                                                                       \\
                                       \toprule
                                       & Original Model            & 0.5                                 & 0.033                            & 0.425                                 & 0.092                               & 0.1                                        & 0.164                                       & 0.133                            & 0.207                \\
                                       \midrule
\textbf{Teacher:} & Lowest LP  & 0.77                                & 0.033                            & 0.5                                   & 0.26                                & 0.407                                      & 0.381                                       & 0.1                              & 0.350                \\
\textbf{Qwen2.5-72B}                                       & Middle LP & 0.79                                & 0.1                              & 0.55                                  & 0.25                                & 0.41                                       & 0.416                                       & 0.133                            & 0.378                \\
\textbf{-Instruct}
                                       & Highest LP & 0.778                               & 0.133                            & 0.6                                   & 0.35                                & 0.46                                       & 0.398                                       & 0.167                            & 0.412                \\
                                       \midrule
\textbf{Teacher:}        & Lowest LP  & 0.802                               & 0.133                            & 0.525                                 & 0.213                               & 0.331                                      & 0.436                                       & 0.2                              & 0.377           \\
\textbf{Gemma3-27B}                                         & Middle LP  & 0.792                               & 0.1                              & 0.575                                 & 0.246                               & 0.312                                      & 0.45                                        & 0.367                            & 0.406                \\
\textbf{-IT}  
                                       & Highest LP & 0.816                               & 0.167                            & 0.625                                 & 0.25                                & 0.387                                      & 0.455                                       & 0.433                            & 0.448   
                                       \\ 
                                       \bottomrule
\end{tabular}
}
\caption{GALP selection within single teacher (short responses). Highest LP consistently outperforms lowest LP within each teacher, confirming that GALP works in this controlled setting. However, rankings do not transfer across teachers (e.g., lowest LP from Qwen2.5-72B gives 0.292 while lowest LP from Gemma3-27B gives 0.279 for different underlying reasons).}\label{tab:within_teacher_non_reasoning_results}
\end{table*}

\begin{figure}[h!]
    \centering
    \begin{minipage}{0.48\textwidth}
        \centering
        \includegraphics[width=\textwidth]{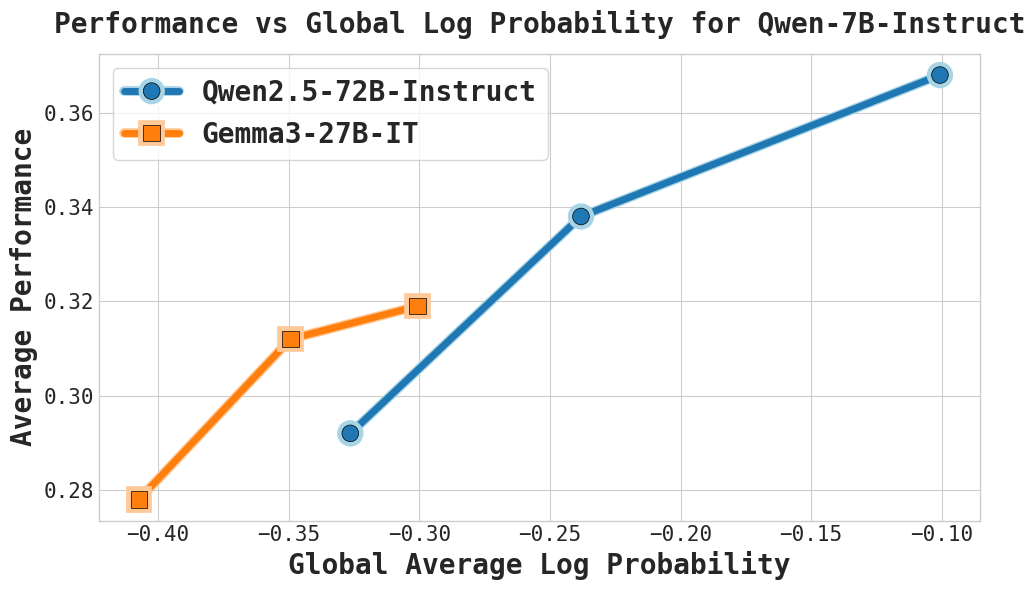}
    \end{minipage}
    \hfill
    \begin{minipage}{0.48\textwidth}
        \centering
        \includegraphics[width=\textwidth]{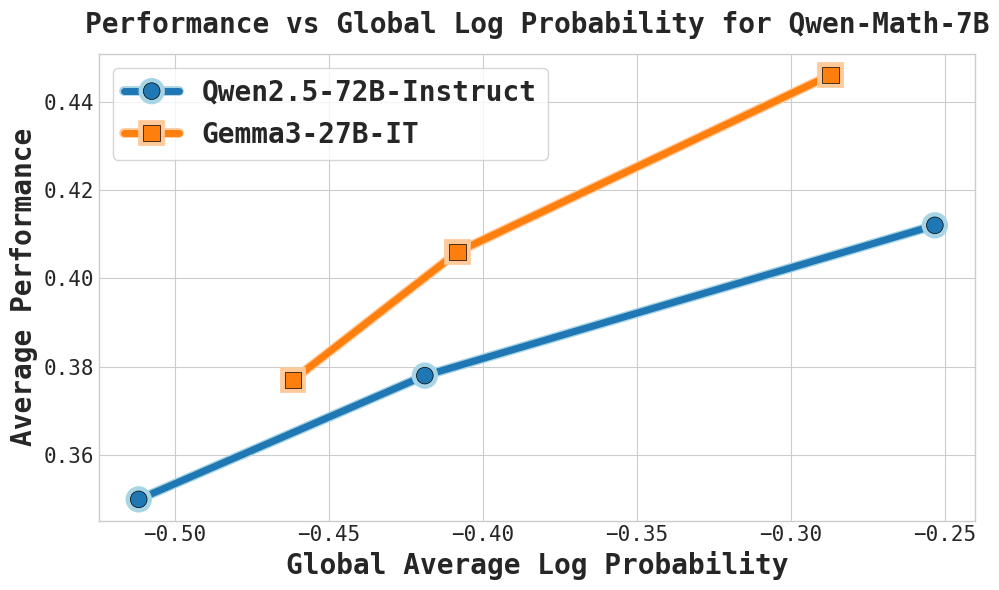}
    \end{minipage}
    \caption{Within-teacher correlation between GALP and performance. Within each teacher (same color), higher global LP correlates with better performance. However, the lines do not align across teachers, indicating GALP cannot reliably compare responses from different sources.}
    \label{fig:global_logprobs_performance_side_by_side}
\end{figure}

\paragraph{Observations.} Table~\ref{tab:within_teacher_non_reasoning_results} shows that within each teacher, highest-GALP selection consistently outperforms lowest-GALP selection. Figure~\ref{fig:global_logprobs_performance_side_by_side} confirms a positive within-teacher correlation. However, the cross-teacher rankings do not align: responses with similar global LP from different teachers yield different performance. This foreshadows the failure mode we examine in the main text, where mixing diverse teachers causes GALP to break down entirely.

\subsection{Extended Results for Within-teacher Sanity Check for GALP}

We provide a benchmark breakdown of results for the GALP selection within a single teacher in Table~\ref{tab:within_teacher_galp_full}.

\begin{table*}[t]
\resizebox{\textwidth}{!}{%
\begin{tabular}{llrrrrrrrr}
                                       &   & \textbf{MATH}             & \multicolumn{1}{l}{\textbf{AIME25}} & \multicolumn{1}{l}{\textbf{AMC}} & \multicolumn{1}{l}{\textbf{MINERVA}} & \multicolumn{1}{l}{\textbf{KAOYAN}} & \multicolumn{1}{l}{\textbf{OLYMPIADB}} & \multicolumn{1}{l}{\textbf{CN\_MATH24}} & \multicolumn{1}{l}{\textbf{AVG}}  \\ 
                                       \toprule
                                       & \multicolumn{9}{c}{\textbf{Student: Qwen2.5-7B-Instruct}} \\ \toprule
                                       & Original Model           & 0.752 & 0.167 & 0.500 & 0.268 & 0.216 & 0.404 & 0.167 & 0.353 \\
                                       \midrule
\textbf{Teacher:} & Lowest GALP  & 0.678 & 0.100 & 0.300 & 0.224 & 0.296 & 0.314 & 0.133 & 0.292 \\
\textbf{Qwen2.5-72B} & Middle GALP  & 0.710 & 0.100 & 0.425 & 0.257 & 0.336 & 0.339 & 0.200 & 0.338 \\
\textbf{-Instruct}   & Highest GALP & 0.744 & 0.133 & 0.500 & 0.252 & 0.391 & 0.391 & 0.167 & 0.368 \\
                                       \midrule
\textbf{Teacher:} & Lowest GALP  & 0.686 & 0.100 & 0.400 & 0.246 & 0.286 & 0.324 & 0.200 & 0.320 \\
\textbf{QwQ-32B}  & Middle GALP  & 0.710 & 0.167 & 0.425 & 0.272 & 0.336 & 0.333 & 0.200 & 0.349 \\
                  & Highest GALP & 0.732 & 0.167 & 0.475 & 0.279 & 0.412 & 0.382 & 0.233 & 0.382 \\
                                       \bottomrule
\end{tabular}
}
\caption{Within-teacher sanity check for GALP. For each teacher, responses are partitioned into terciles by the pre-SFT student's global average log probability (Eq.~\ref{eqn:galp}). Fine-tuning on the highest-GALP tercile yields the strongest average downstream performance.}
\label{tab:within_teacher_galp_full}
\end{table*}

\subsection{Generalizability Experiments}
\label{app:gpqa_lcb}
To explicitly test this generalizability, we have since conducted additional experiments in general science. For the models trained on math data from our paper, we computed performance the GPQA-Diamond benchmark, which tests expert-level reasoning across biology, physics, and chemistry. We obtained the following results:

\begin{table}[h!]
\centering
\caption{GPQA-Diamond Benchmark Results}
\label{tab:gpqa_results}
\resizebox{0.5\columnwidth}{!}{%
\begin{tabular}{@{}lc@{}}
\toprule
\textbf{Model / Method} & \textbf{GPQA-Diamond (pass@1)} \\ \midrule
Original Qwen2.5-32B-Instruct & 0.551 \\
All 3 Teachers’ Responses & 0.439 \\
LIMO-32B & 0.626 \\
Sky-T1-32B-Preview & 0.566 \\
OpenThinker2-32B & 0.646 \\
Global Highest (GRAPE) & 0.611 \\
Local Highest (Ours) & 0.702 \\ \bottomrule
\end{tabular}%
}
\end{table}

Notably, our LALP selection method not only surpasses the Global Highest baseline but also outperforms other state-of-the-art models. This is particularly significant as these other models were trained on substantially larger and more diverse long reasoning datasets, underscoring the efficiency and effectiveness of our data curation technique.

To further demonstrate the generalizability of our approach, we extended our evaluation to code reasoning. We generated responses for 5,000 prompts from the OpenCodeReasoning and LeetCode datasets and used them to fine-tune the Qwen2.5-32B-Instruct model. The models performance were then evaluated on the LiveCodeBench v2 benchmark.

\begin{table}[h!]
\centering
\label{tab:livecodebench_results}
\resizebox{\columnwidth}{!}{%
\begin{tabular}{@{}lccc@{}}
\toprule
\textbf{Method} & \textbf{LiveCodeBench-easy} & \textbf{LiveCodeBench-medium} & \textbf{LiveCodeBench-hard} \\ \midrule
Original Qwen2.5-32B-Instruct & 0.890 & 0.471 & 0.114 \\
Global Highest (GRAPE) & 0.845 & 0.588 & 0.232 \\
Local Highest (Ours) & 0.874 & 0.633 & 0.261 \\ \bottomrule
\end{tabular}%
}
\caption{LiveCodeBench v2 Benchmark Results}
\end{table}

As the results indicate, the student model trained on data selected via LALP consistently outperforms the one trained using the global log-probability baseline across the medium and hard difficulty tiers.

These additional results from the scientific and coding domains strongly suggest that the core principle of LALP---step-level standalone justifiability---is not confined to mathematics. The method's effectiveness in identifying high-quality reasoning data appears to generalize to other domains that require complex, step-by-step inference. We will include these findings in the paper to provide a more comprehensive evaluation of our method's applicability.

\subsection{Ablation: Llama-3.1-8B-Instruct} 
\label{app:llama3}

We provide results on another student model to show the generalizability of our method in Table~\ref{tab:llama}.

\begin{table}[h!]
\resizebox{0.99\textwidth}{!}{%
\begin{tabular}{lllllllll}
\textbf{}                & \multicolumn{1}{c}{\textbf{MATH}} & \multicolumn{1}{c}{\textbf{AIME25}} & \multicolumn{1}{c}{\textbf{AMC}} & \multicolumn{1}{c}{\textbf{MINERVA}} & \multicolumn{1}{c}{\textbf{KAOYAN}} & \multicolumn{1}{c}{\textbf{OLYMPIAD}} & \multicolumn{1}{c}{\textbf{CN\_MATH24}} & \multicolumn{1}{c}{\textbf{GPQA}} \\

\toprule
\multicolumn{9}{c}{{\textbf{Student: Llama-3.1-8B-Instruct}}}                                                                                                                                                                                                                                                                               \\
\toprule
\textbf{Original Model} & 0.726 	&0.0 &	0.45 &	0.316 	&0.216 &	0.361 	&0.3 &	0.656      \\
\midrule
\textbf{All 3 Teachers}  & 0.794 &  0.167 &		0.725 	&	0.335 	&	0.432 	&	0.51 	&	0.2 &		0.878   \\
\textbf{Global Highest}  &    0.796 	&0.133 	&0.625 	&0.371 	&0.437	&0.48 	&0.2 	&0.833    \\
\textbf{Local Highest (Ours)}   &   0.814 	&0.167	&0.8   	&0.368 &	0.467 &	 0.49 	&0.233 	& 0.883
                                       \\ 
                                       \bottomrule       
\end{tabular}
}
\caption{Performance of Llama-3.1-8B-Instruct student models on LIMO prompts when fine-tuned with responses from different selection strategies. The log probabilities (LP) are the global and local average log probabilities of the student model with non-greedy decoding with temperature 0.6, top-p 0.95 and over 8 samples, pass@8.}\label{tab:llama}
\vspace{-1em}
\end{table}

\subsection{Cross Teacher Selection: Performance with Non-Greedy Decoding}
\label{app:non_greedy_decoding}

We provide results on cross teacher selection with non-greedy decoding in Table~\ref{tab:data_selection_results_acrosss_teacher_models greedy}.

\begin{table}[h!]
\resizebox{0.99\textwidth}{!}{%
\begin{tabular}{lllllllll}
\textbf{}                & \multicolumn{1}{c}{\textbf{MATH}} & \multicolumn{1}{c}{\textbf{AIME25}} & \multicolumn{1}{c}{\textbf{AMC}} & \multicolumn{1}{c}{\textbf{MINERVA}} & \multicolumn{1}{c}{\textbf{KAOYAN}} & \multicolumn{1}{c}{\textbf{OLYMPIAD}} & \multicolumn{1}{c}{\textbf{CN\_MATH24}} & \multicolumn{1}{c}{\textbf{AVG}} \\

\toprule
\multicolumn{9}{c}{{\textbf{Student: Qwen2.5-32B-Instruct}}}                                                                                                                                                                                                                                                                               \\
\toprule
\textbf{Original Model} & 0.826 	&0.121 &	0.715 &	0.295 	&0.416 &	0.461 	&0.237 &	0.439      \\
\midrule
\textbf{All 3 Teachers}          &   0.835 &		0.400 &		0.887 	&	0.335 	&	0.578 	&	0.559 	&	0.583 &		0.597   \\
\textbf{Global Highest}  &    0.862 	&0.442 	&0.903 	&0.338 	&0.629 	&0.634 	&0.721 	&0.647    \\
\textbf{Local Highest (Ours)}   &   0.911 	&0.662 	&0.969 	&0.361 &	0.658 &	0.675 	&0.850 	& \textbf{0.727}
                                       \\ 
                                       \bottomrule       
\end{tabular}
}
\caption{Performance of Qwen2.5-7B-Instruct and Qwen2.5-32B-Instruct student models on LIMO prompts when fine-tuned with responses from different selection strategies. The log probabilities (LP) are the global and local average log probabilities of the student model with non-greedy decoding with temperature 0.6, top-p 0.95 and over 8 samples.}\label{tab:data_selection_results_acrosss_teacher_models greedy}
\vspace{-1em}
\end{table}

% \newpage
\subsection{Performance Comparison with Other SOTA Qwen-32B Models}
\label{app:sota_models}

We compare the performance of our model against several strong open-source implementations that also fine-tune the Qwen-32B-Instruct student model on comparable or larger datasets. LIMO-32B-V1\citep{ye2025limo} (available at \url{https://huggingface.co/GAIR/LIMO}) is trained on the LIMO prompt set using responses exclusively from the DeepSeek-R1 teacher. Sky-T1-32B-Preview\citep{sky_t1_2025} (available at \url{https://huggingface.co/NovaSky-AI/Sky-T1-32B-Preview}) is trained on a 17K example dataset~(\url{https://huggingface.co/datasets/NovaSky-AI/Sky-T1_data_17k}) generated using the QWQ-32B model. OpenThinker2-32B\citep{openthoughts} (available at \url{https://huggingface.co/open-thoughts/OpenThinker2-32B}) is trained on a substantially larger dataset of 1.04M samples(\url{https://huggingface.co/datasets/open-thoughts/OpenThoughts2-1M}), also generated using DeepSeek-R1 as the teacher. A detailed comparison of the results is provided in Table~\ref{tab:model 32 across sota methods}.

\begin{table}[h!]
\resizebox{0.99\textwidth}{!}{%
\begin{tabular}{lllllllllll}
\textbf{}                & \multicolumn{1}{c}{\textbf{MATH}} & \multicolumn{1}{c}{\textbf{AIME25}} & \multicolumn{1}{c}{\textbf{AMC}} & \multicolumn{1}{c}{\textbf{MINERVA}} & \multicolumn{1}{c}{\textbf{KAOYAN}} & \multicolumn{1}{c}{\textbf{OLYMPIAD}} & \multicolumn{1}{c}{\textbf{CN\_MATH24}} & \multicolumn{1}{c}{\textbf{AVG}}    & \multicolumn{1}{c}{\textbf{GPQA}}                                                                                                    \\
\toprule
\multicolumn{9}{c}{{\textbf{Student: Qwen2.5-32B-Instruct}}}   \\
\toprule
\textbf{Original Model} & 0.824        & 0.133          & 0.700       & 0.298             & 0.422        & 0.471          & 0.233          & 0.445  & 0.551      \\
\midrule
\textbf{Global Highest}  &    0.876     &  0.433       &   0.825     &   0.331       &       0.592        &   0.636       &      0.733      &     0.632 &     0.611    \\
\textbf{LIMO-32B}  &    0.896     &  0.433       &   0.925     &   0.346       &       0.618        &   0.630       &      0.800      &     0.664  &     0.626    \\
\textbf{Sky-T1-32B-Preview}  &    0.876     &  0.200      &   0.750     &   0.301      &       0.558        &   0.507       &      0.533      &     0.532 &     0.566    \\
\textbf{OpenThinker2-32B}  &    0.922     &  0.567       &   0.900     &   0.324       &       0.648        &   0.640       &      0.833      &     0.691&     0.646    \\
\textbf{Local Highest (Ours)}   &   0.902     &   0.667     &     1.000    &     0.353       &     0.653         &   0.673      &      0.833       &   \textbf{0.726} &      0.694   
                                       \\ 
                                       \bottomrule       
\end{tabular}
}
\caption{A performance comparison of our model with other open-source SOTA models fine tuned on the Qwen2.5-32B-Instruct student model.}
\label{tab:model 32 across sota methods}
\vspace{-1em}
\end{table}

%%%%%%%%%%%%%%%%%%%%%%%%%%%%%%%%%%%%%%%%%%%%%%%%%%%%%%%%%%%%%%%%%%%%%%%%%%%%%%%
%%%%%%%%%%%%%%%%%%%%%%%%%%%%%%%%%%%%%%%%%%%%%%%%%%%%%%%%%%%%%%%%%%%%%%%%%%%%%%%

\end{document}